\documentclass[journal]{IEEEtran}

\usepackage{times}
\usepackage{epsfig}
\usepackage{graphicx}
\usepackage{amsmath}
\usepackage{amssymb}
\usepackage{enumitem}
\usepackage{tabu}
\usepackage{booktabs}
\usepackage{multirow}
\usepackage{array}
\usepackage{adjustbox}
\usepackage{xcolor}
\usepackage{cite}
\usepackage{url}

\def\tabvspace{{\vspace{-2mm}}}

\def\figvspaceloose{{\vspace{-2mm}}}

\begin{document}

\title{Towards a Unified Approach to Single Image Deraining and Dehazing}


\author{Xiaohong~Liu,~\IEEEmembership{Graduate~Student~Member,~IEEE,}
	Yongrui Ma,~
	Zhihao Shi,~
	Linhui Dai, and~Jun~Chen,~\IEEEmembership{Senior Member,~IEEE}
	\thanks{X. Liu, Y. Ma, Z. Shi, L. Dai, and J. Chen (Corresponding Author) are with the Department of Electrical and Computer Engineering, McMaster University, Hamilton, ON L8S 4K1, Canada (e-mail: \{liux173, may85, shiz31, dail5, chenjun\}@mcmaster.ca). This work was
		supported in part by the Natural Sciences and Engineering Research Council
		of Canada through a Discovery Grant.}}

\maketitle

\begin{abstract}
	We develop a new physical model for the rain effect and show that the well-known atmosphere scattering model (ASM) for the haze effect naturally emerges as its homogeneous continuous limit. Via depth-aware fusion of multi-layer rain streaks according to the camera imaging mechanism, the new model can better capture the sophisticated non-deterministic degradation patterns commonly seen in real rainy images. We also propose a Densely Scale-Connected Attentive Network (DSCAN) that is suitable for both deraining and dehazing tasks. Our design alleviates the bottleneck issue existent in conventional multi-scale networks and enables more effective information exchange and aggregation. Extensive experimental results demonstrate that the proposed DSCAN is able to deliver superior derained/dehazed results on both synthetic and real images as compared to the state-of-the-art. Moreover, it is shown that for our DSCAN, the synthetic dataset built using the new physical model yields better generalization performance on real images in comparison with the existing datasets based on over-simplified models.
\end{abstract}


\section{Introduction}

\IEEEPARstart{S}{ingle} image deraining/dehazing aims to recover a clear image from a rainy/hazy version. Deraining and dehazing techniques can be leveraged 
to alleviate image degradation caused by adverse weather conditions, which is crucial for developing robust surveillance and autonomous driving systems~\cite{bijelic2018benchmarking,kong2017millimeter}. Recent advances in deep learning provide a significant boost to this line of research as evidenced by a rapidly growing body of literature.

\begin{figure}[t]
	\centering
	\begin{minipage}[b]{0.49\linewidth}
		\centering
		\includegraphics[width=\linewidth]{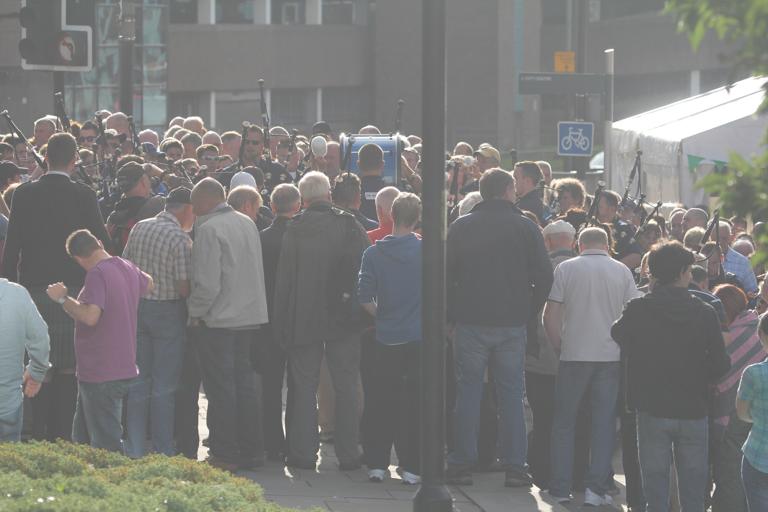}
		\scriptsize{(a) Hazy image}
	\end{minipage}
	\begin{minipage}[b]{0.49\linewidth}
		\centering
		\includegraphics[width=\linewidth]{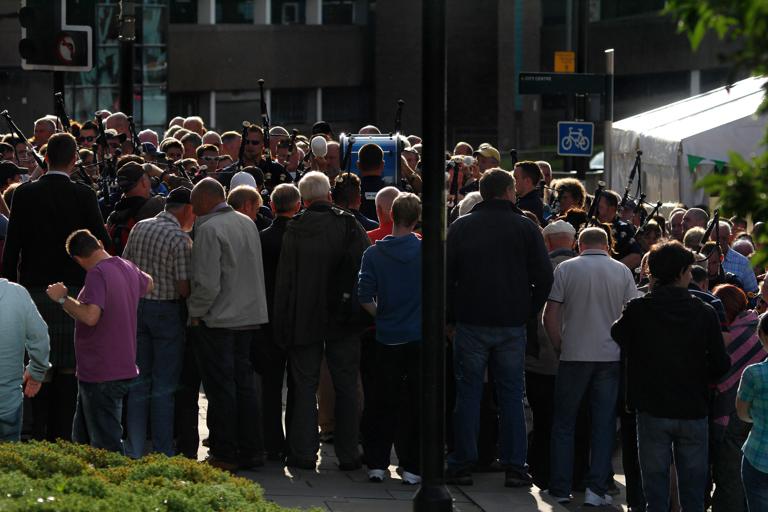}
		\scriptsize{(b) Our dehazed result}
	\end{minipage}
	\begin{minipage}[b]{0.49\linewidth}
		\centering
		\includegraphics[width=\linewidth]{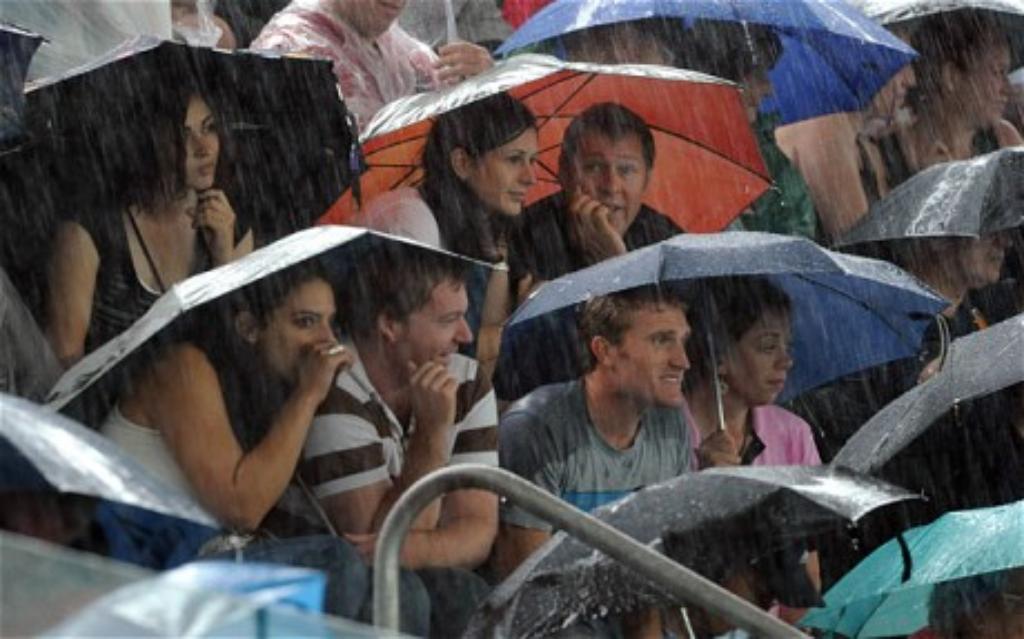}
		\scriptsize{(c) Rainy image}
	\end{minipage}		
	\begin{minipage}[b]{0.49\linewidth}
		\centering
		\includegraphics[width=\linewidth]{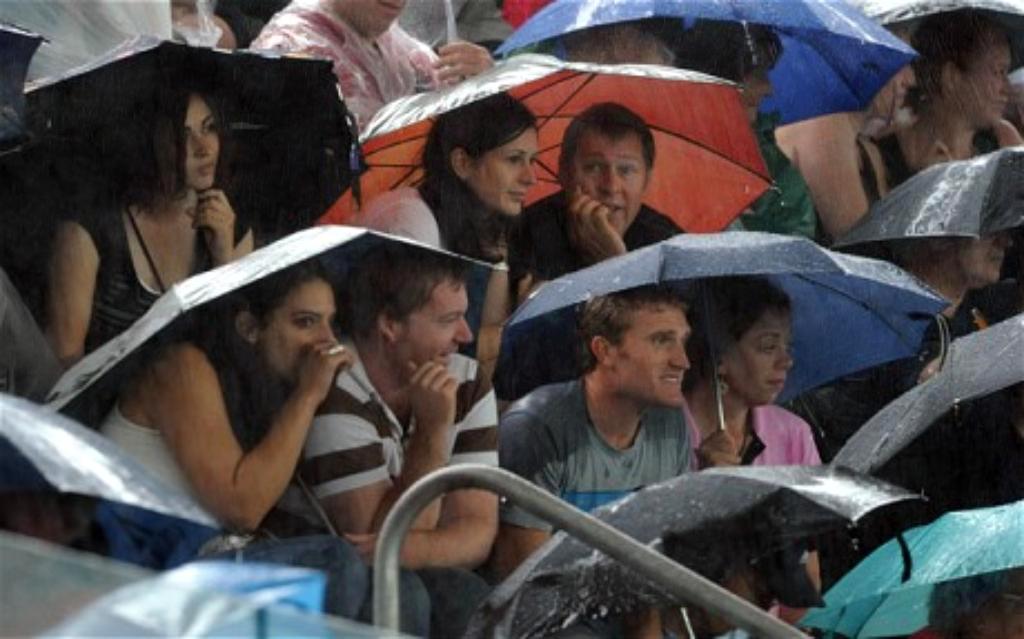}
		\scriptsize{(d) Our derained result}
	\end{minipage}	
	\caption{Examples of deraining and dehazing with the proposed DSCAN applied on real images.}
	\label{fig:introduction}
	\figvspaceloose
\end{figure}

Somewhat surprisingly, although deraining and dehazing share many similarities, they are often treated separately~\cite{fu2017removing,zhang2018density,ren2019progressive,iebmmscnn01,wang2019spatial,iebmaod01,iebmgated01}. This situation is arguably a result of the fact that there is no unified physical model for rain and haze effects. It is known that  the standard atmosphere scattering model (ASM)  \cite{mccartney1976optics,narasimhan2000chromatic,narasimhan2002vision} for the haze effect is inadequate for capturing the sophisticated non-deterministic degradation patterns commonly seen in real rainy images. Even worse, the homogeneous infinitesimal analysis used in the derivation of ASM appears to inherently unsuitable for rain streaks due to their relative sparsity and non-negligible size.
The existing rain models are all heuristic in nature, and to the best of our knowledge, the counterpart of ASM for the rain effect is not yet available in the literature.
The lack of a unified model for rain and haze effects also makes it difficult to interpret some empirical results (e.g., the benefit of joint learning for rain/haze removal~\cite{yang2017deep,li2019single}). Data-driven deraining and dehazing methods typically need to be trained on large datasets in a supervised fashion. However, collecting a large number of real rainy/hazy images and their clear version is a formidable task, and the alternative data generation approach based CycleGAN~\cite{zhu2017unpaired} is not guaranteed to deliver realistic-looking images due to the instability issue~\cite{goodfellow2014generative}. Therefore, one often resorts to physical-model-based synthetic datasets for supervised training. This sets a high standard on the accuracy of the adopted physical model since model mismatch may lead to poor generalization performance on real images. It is fair to say that the existing rain models do not quite   meet the requirement. 

On the neural network design side, the multi-scale structure has been a popular choice for single image deraining and dehazing. This structure allows for simultaneous exploitation low-level features (which retain information relevant to texture details) and high-level features (which are robust against image degradation),  making it particularly suitable for image restoration tasks~\cite{shen2018deep,chen2018learning}. However, the conventional multi-scale structure tends to suffer from the bottleneck issue due to limited information exchange across different scales and inflexible fusion mechanisms.

In this work, we make some progress towards a unified approach to deraining and dehazing. Our main contributions are summarized as follows: 1) We develop a new physical model for the rain effect based depth-aware fusion of multi-layer rain streaks according to the camera imaging mechanism, and show that the well-known ASM for the haze effect naturally emerges as its homogeneous continuous limit. 2) We propose a Densely Scale-Connected Attentive Network (DSCAN) that is suitable for both deraining and dehazing tasks;
our design alleviates the bottleneck issue existent in conventional multi-scale networks and enables more effective information exchange and aggregation. 3) We conduct extensive experiments to demonstrate the competitive performance of the proposed DSCAN and validate our unified physical model for rain and haze effects.  Fig.~\ref{fig:introduction} shows some examples of deraining and dehazing with the proposed DSCAN applied on real images.

\section{Related Work}
In this section, we provide a brief review of some representative deraining and dehazing methods and the existing rain/haze models.


\subsection{Deraining and Dehazing Methods}

\subsubsection{Dehazing} In contrast to early research \cite{multiimagepolar01,multiimagepolar02,scatteringfn02,multiimageweather02} which typically assumes the availability of certain side information to aid haze removal, recent works have paid more attention to the more challenging single image dehazing problem. Essentially all the existing single image dehazing methods depend on ASM 
(to be described in Section~\ref{sec:physical models}) in one way or another. Most works \cite{iebmcontrast01,iebmalbedo01,iebmdcp01,iebmrf01,iebmlinear01,iebmdehazenet01,iebmmscnn01,iebmaod01} make explicit use of ASM  by reducing the dehazing problem to the estimation of the unknown parameters in ASM. Some other works \cite{iebmgated01} bypass ASM 
in algorithm design, but still rely on ASM-based datasets for supervised training, thereby implicitly leveraging the knowledge of ASM.



\subsubsection{Deraining} The existing deraining methods can be roughly divided into three categories~\cite{li2019single}: multi-frame-based methods~\cite{ren2017video,santhaseelan2015utilizing,jiang2017novel}, prior-based methods~\cite{barnum2010analysis,zheng2013single,li2016rain}, and deep-learning-based methods~\cite{yang2017deep,fu2017removing,zhang2018density,li2018recurrent,wang2019spatial,ren2019progressive}. Here we focus on the last category as it is most relevant to the present work. The deep-learning-based approach to image deraining was initiated by \cite{fu2017removing}. Some improvements were made via the inclusion of rain-density estimation  \cite{zhang2018density} and attention mechanism \cite{wang2019spatial}. Recurrent processing was adopted in \cite{yang2017deep},~\cite{li2018recurrent}, and~\cite{ren2019progressive} to realize image deraining with compact network design. In addition, \cite{qian2018attentive,meng2018removal,li2019heavy,zhang2019image} proposed the use of generative adversarial network (GAN) to improve the perceptual quality of derained results.


\subsection{Physical Models}\label{sec:physical models}
\subsubsection{Haze} The atmosphere scattering model (ASM) \cite{mccartney1976optics,narasimhan2000chromatic,narasimhan2002vision} provides a reasonable approximation of the haze effect and can be expressed as 
\begin{align}
	H^{(c)}(x)=B^{(c)}(x)t(x)+A(1-t(x)), \label{eq:asm}
\end{align}
where $H^{(c)}(x)$ ($B^{(c)}(x)$) is the intensity of pixel $x$ in color channel $c\in\{r,g,b\}$ of  hazy (clear) image $H$ ($B$), $t(x)$ is the transmission map, and $A$ is the global atmospheric light. Moreover, we have $t(x) = e^{-\beta d(x)}$ with $\beta$ and $d(x)$ representing the atmosphere scattering coefficient and the scene depth respectively.


\subsubsection{Rain Streak} A simple way to model the rain effect is superimposing a rain-streak layer on the clear image~\cite{fu2017removing, eigen2013restoring, fu2017clearing, zhang2018density, li2016rain}:
\begin{align} \label{equ:basic}
	R^{(c)}(x) = B^{(c)}(x) + S(x),
\end{align}
where $R^{(c)}(x)$ stands for the intensity of pixel $x$ in color channel $c$ of rainy image $R$, and $S(x)$ is the intensity of pixel $x$ of rain-streak layer $S$. The multi-layer version \cite{li2018recurrent} of (\ref{equ:basic}) is given by 
\begin{align}\label{eq:multilayer}
	R^{(c)}(x) = B^{(c)}(x) +  \sum\nolimits_{i=1}^{L}S_{i}(x),
\end{align}
where $S_i(x)$ is the intensity of pixel $x$ of the $i$-th rain-streak layer $S_i$, and $L$ denotes the total number of layers. Note that (\ref{equ:basic}) and (\ref{eq:multilayer}) are mathematically equivalent  since one can simply interpret the superposition of multiple rain-streak layers as a single layer.  However, in practice, 
it is often assumed that the rain streaks in the same layer are of similar characteristics, and as a consequence, multiple layers are needed to simulate complex rain effects.





\subsubsection{Rain Streak and Haze} 
It is known that, in heavy rain, visual degradation of distant scenes  is similar to that seen in the hazy atmospheric condition. This is known as the rain veiling effect, which is attributed to light scattering caused by rain accumulation. Some attempts \cite{li2019single,li2018recurrent,li2019heavy,yang2017deep} have been made to simulate this effect by integrating ASM with the rain-streak-based model.
There are two different integration approaches, depending on the superposition order of rain and haze effects.
The haze-first approach~\cite{li2019single} can be formulated as
\begin{align} \label{equ:withasm1}
	R^{(c)}(x) = B^{(c)}(x)t(x) + A(1 - t(x)) + S(x).
\end{align}
The above model can be further refined by introducing multiple rain-streak layers with potentially different brightness \cite{li2018recurrent}. In contrast, the rain-first approach \cite{li2019heavy} superimposes rain-streak layers on the clear image before introducing the haze effect, and the resulting model can be expressed as 
\begin{align}
	R^{(c)}(x) = \left(B^{(c)}(x) + \sum_{i=1}^{L}S_{i}(x)\right) t(x) + A(1-t(x)).
\end{align}
This model can also be refined in various ways (see, e.g., \cite{yang2017deep} for a variant of this model which can better simulate inhomogeneously distributed rain streaks 
through the use of a region-dependent binary map to indicate the location of visible rain streaks).

However, all the aforementioned rain models are heuristic in nature and lack solid physical justifications. The way that the rain-streak layers are used to simulate the rain effect is not based on the camera imaging mechanism. In particular, the scene depth information is not systemically used (as far as the construction of rain-streak layers is concerned), and light scattering caused by rain accumulation is not properly addressed. Moreover, it is conceptually unsatisfactory to leverage ASM to simulate the rain veiling effect. Indeed, a good rain model should be able to exhibit such effect without having it explicitly built in.










\section{Our Unified Physical Model}
In this work, we develop a new physical model for the rain effect according to the camera imaging mechanism. This new model naturally gives rise to the rain veiling effect without relying on ASM. In fact, we show that ASM can be viewed as its homogeneous continuous limit. Our unified model also sheds light on the empirical observation in \cite{li2019heavy} that pre-training on hazy images helps improve the quality of derained results.


The construction of our unified model takes two steps: rain streak generation and rain effect fusion. It will be seen that the underlying idea is quite general and, with some adjustment, can be used to construct physical models for other atmospheric conditions (e.g., snow).

\subsection{Rain Streak Generation}

\begin{figure}[t]
	\centering
	\includegraphics[width=\linewidth]{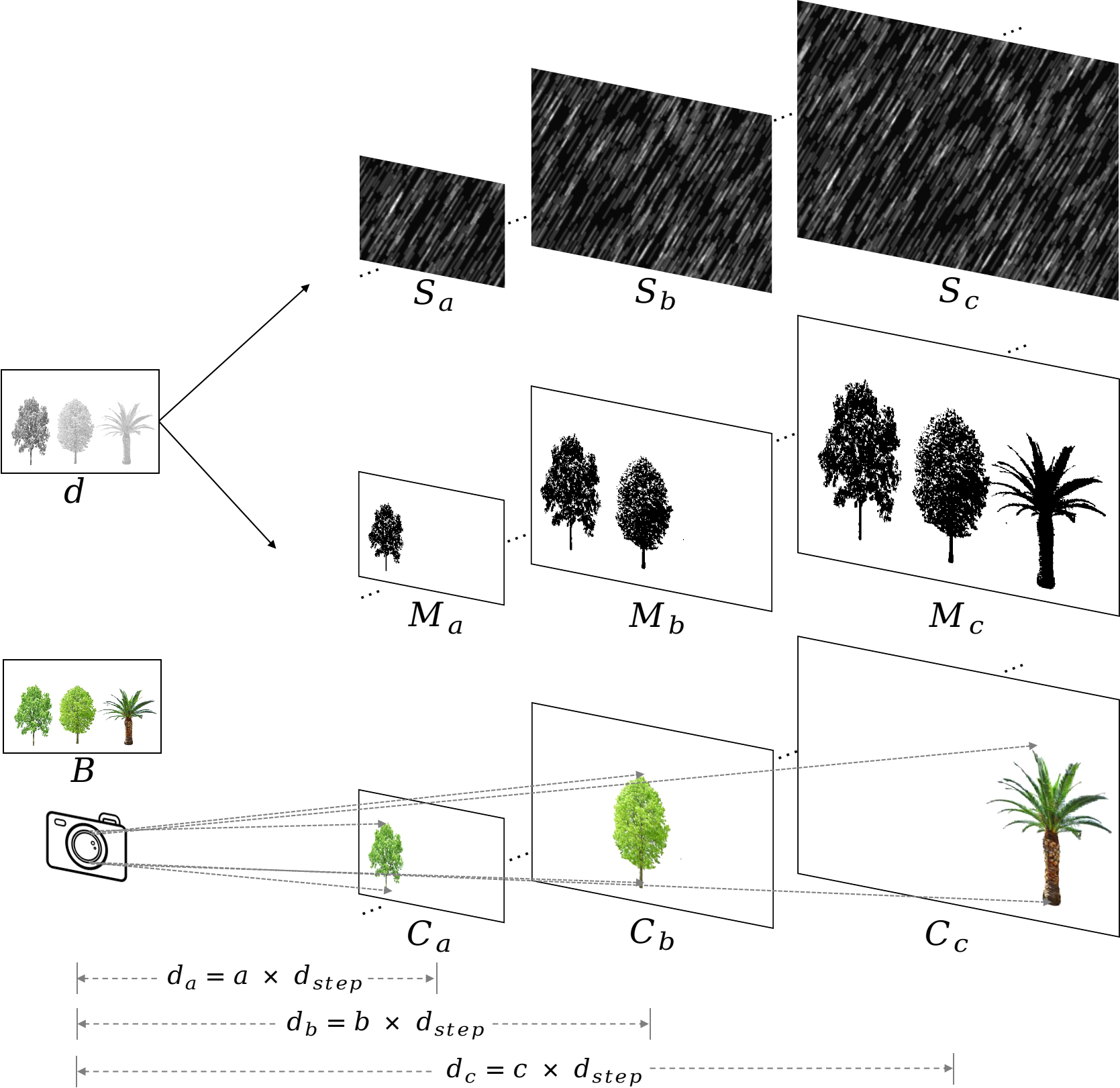}
	\caption{An illustration of the proposed physical model. Here $B$ is a clear image and $d$ is its depth map. We partition $B$ into a number of slices with each slice containing objects located in a disjoint depth interval of length $d_{step}$. These slides are reshaped according to their respective scale factors $s$. A rain-streak layer, accompanied by a rain mask, is generated for each slice. For simplicity, here we only show three slices each containing a tree: $C_a$ (with scene depth $d_a$), $C_b$ (with scene depth $d_b$), and $C_c$ (with scene depth $d_c$), where $a=d_a/d_{step}$, $b=d_b/d_{step}$, and $c=d_c/d_{step}$. The corresponding rain-streak layers $S_a$, $S_b$, and $S_c$ as well as the associated rain masks $M_a$, $M_b$, and $M_c$ are also depicted. Note that an element of rain mask is set to 0 (black) if the scene depth of the corresponding pixel in $B$ is less than or equal to the depth of this mask.
		It is also worth noting that each pixel in $B$ corresponds to a patch in the reshaped slice, and the patch size increases with the scene depth of the slice.}
	\label{fig:rain_streak}
	\figvspaceloose
\end{figure}

We first partition the given clear image $B$ into a total of $k$ slices according to the scene depth $d(x)$, where $k$ is determined by the maximum scene depth $d_{max}$ and the slice step $d_{step}$.
As a result, the objects with different scene depths appear in their associated image slices (e.g., the trees in Fig.~\ref{fig:rain_streak} are shown in different slices according to their respective distances to the camera lens). The produced slices are then linearly reshaped according to their respective scale factors $s$. Since the object distance $u$ is typically much greater than the camera focal length $f$ \cite{cai2019toward}, we have
$s = \frac{h_1}{h_2} \approx \frac{u}{f}$,
where $h_1$ and $h_2$ denote respectively the actual height of the object and the depicted height in the image. For each slice, we generate a rain-streak layer by following the strategy in \cite{zhang2018density} (the rain-streak layer associated with the $i$-th slice is denoted by $S_i$). Specifically, we produce a set of random points using a Poisson point process and apply a layer-specific rain kernel on each point via a 2D convolutional operation. The physical attributes of the generated rain streaks (i.e., density, length and direction) can be controlled by adjusting the parameters of  Poisson point process and rain kernels.

\begin{figure*}[t]
	\centering
	\includegraphics[width=\linewidth]{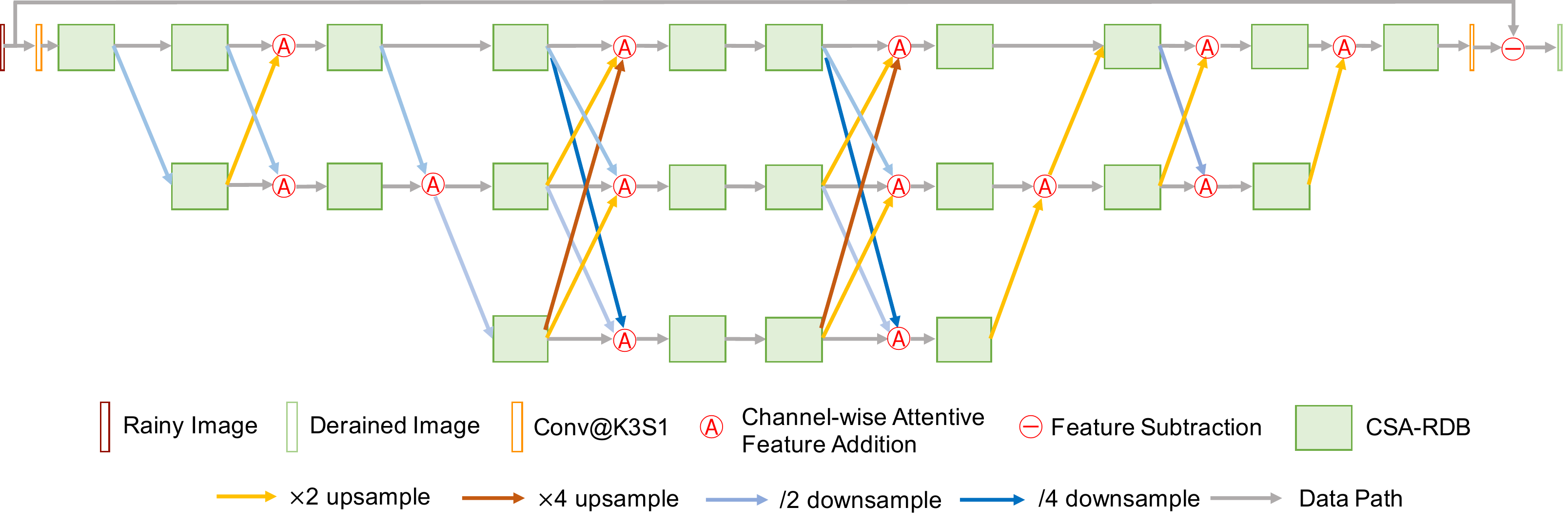}
	\caption{The overall architecture of the proposed DSCAN. Conv@K$n$S$m$ denotes a $n \times n$ convolution with stride $m$.}
	\label{fig:DSCAN}
	\figvspaceloose
\end{figure*}

\subsection{Rain Effect Fusion}
Since a rain streak can be observed only if there does not exist any obstruction ahead, a rain mask is constructed to indicate the occluded region of each rain-streak layer. Let $M_i$ denote the rain mask associated with the $i$-th rain-streak layer $S_i$. Note that $M_i$ and $S_i$ (as well as the reshaped $i$-th slice) are of the same size. We use $M_i(x)$ ($S_i(x)$) to represent the patch in $M_i$ ($S_i$) that is associated with pixel $x$ in the given clear image $B$, and set the elements of $M_i(x)$ to $0$ for $x$ in the occluded region (i.e., when $d(x)$ is less than or equal to the depth of $S_i$) and to $1$ otherwise. For the masked rain-streak patch $S_i(x) \odot M_i(x)$ (with $\odot$ being element-wise multiplication), we define the rain intensity $q_i(x)$ as
\begin{align}
	q_i(x)=F^{box}_i(S_i(x)\odot M_i(x)),
	\label{eq:qi}
\end{align}
where $F^{box}_i$ denotes the box filter. Note that $q_i(x)$ can be interpreted as the fraction of area of the masked patch $S_i(x) \odot M_i(x)$  being covered by rain streaks.

It is reasonable to assume that the light intensity (with respect to pixel $x$) is reduced by $\alpha q_i(x)$ (in the fractional sense) due to the $i$-th rain-streak layer, where $\alpha$ is the attenuation ratio used to quantify the extent by which rain streaks block the incoming light. Therefore, by considering the cumulative effect of all rain-streak layers, we have
\begin{equation}
	B^{(c)}_d(x)=B^{(c)}(x)\prod_{i=1}^{k(x)}(1-\alpha q_i(x)),
\end{equation}
where $B^{(c)}_d(x)$ and $B^{(c)}(x)$ are respectively the background light intensities (measured at the camera end) of pixel $x$ of color channel $c$  with and without rain-streak layers, and $k(x)=\lfloor d(x)/d_{step}\rfloor$. Note that besides attenuating the background light intensities, each rain-streak layer also contributes its own light intensities to the final rainy image. It is reasonable to assume that the light intensity contributed by the $i$-th  rain-streak layer (with respect to pixel $x$) is $A\alpha q_i(x)$ when all the other layers are removed. Now considering the attenuation caused by the first $i-1$ layers, the effective light intensity contribution (with respect to pixel $x$) from the $i$-th  rain-streak layer becomes
\begin{equation}
	Q_i(x)=A\alpha q_i(x) \prod_{j=1}^{i-1}(1-\alpha q_j(x)).
\end{equation}
The accumulated light intensity contribution $S_m(x)$ from all rain-streak layers is given by 
\begin{equation}
	S_m(x)=\sum_{i=1}^{k(x)}Q_i(x).
\end{equation}
Combining $B^{(c)}_d(x)$ and $S_m(x)$ yields our new rain model
\begin{equation}
	\begin{aligned}
		\label{eq:raining_model}
		R^{(c)}(x)&=B^{(c)}(x)\prod_{i=1}^{k(x)}(1-\alpha q_i(x)) \\ & +\sum_{i=1}^{k(x)}A\alpha q_i(x) \prod_{j=1}^{i-1}(1-\alpha q_j(x)).
	\end{aligned}
\end{equation}


Due to the use of Poisson point process in rain streak generation, $q_i(x)$ is in general a random variable. However, since the size of rain-streak patch $S_i(x)$ increases with $i$, it follows by the law of large numbers that $q_i(x)$ is essentially a constant when $i$ is large enough (assuming that $q_i(x)$ only depends on the number of realizations of Poisson point process falling inside $S_i(x)$). That is to say, 
the random rain streak effect (due to the presence/absence of rain streaks) gradually gives way to the deterministic rain veiling effect as the rain-streak layer moves further away from the camera. This observation also suggests that it might be possible to obtain ASM as the homogeneous continuous limit of the new model. Indeed, by setting $q_i(x)=\beta d_{step}/
\alpha$, we can 
rewrite (\ref{eq:raining_model}) as
\begin{equation}
	\begin{aligned}
		\label{eq:forhaze1}
		R^{(c)}(x)&=B^{(c)}(x)(1-\beta d_{step})^{k(x)}\\&+A(1-(1-\beta d_{step})^{k(x)});
	\end{aligned}
\end{equation}
sending $d_{step}\rightarrow 0$ and invoking the fact that $\lim\limits_{x\to \infty} (1 + \frac{a}{x})^{bx}=e^{ab}$ gives
\begin{equation}
	\begin{aligned}
		\label{eq:forhaze2}
		R^{(c)}(x)=B^{(c)}(x)e^{-\beta d(x)}+A(1-e^{-\beta d(x)}),
	\end{aligned}
\end{equation}
which coincides with (\ref{eq:asm}).

In summary, the new model unifies rain and haze effects in at least two aspects: 1) The haze effect can be viewed as an extreme case of the rain effect caused by infinitesimally small rain streaks uniformly distributed in space. 2) Even when the size of rain streaks is not negligible, the model still ensures that the haze effect  arises in the distant area of the scene. Both aspects are captured by the new model in an intrinsic way. This is conceptually satisfactory since there is no need to artificially build the haze effect and the rain effect separately into the model. Moreover, the structure of the new model is largely generic in nature; rain 
characteristics and dynamics enter the model only through $\alpha q_i(x)$. Therefore, it is possible to capture more sophisticated rain effects by suitably adjusting the relevant parameters. It is also worth emphasizing that our model is physically justifiable although for simplicity we choose not to derive it from first principles. In fact, it can be viewed as a discretized version of  ASM and is based on similar physics principles. There are also natural correspondences between the parameters in our model (say, $\alpha q_i(x)$) and those in ASM (say, $\beta$).

\section{Method}

Our unified physical model for rain and haze effects naturally suggests the possibility of designing a single neural network that is suitable for both deraining and dehazing. In this paper, we propose a Densely Scale-Connected Attentive Network (DSCAN) as a candidate solution. The overall architecture of DSCAN is shown in Fig.~\ref{fig:DSCAN}.

The rationale underlying the design of DSCAN is best explained in the general context of multi-scale information exchange and aggregation. The performance of a multi-scale network depends critically on the extent to which its architecture facilitates information flows. For example, in a conventional encoder-decoder network, the information first moves downward in the scale  hierarchy at the encoder end then moves upward at the decoder end; it is clear that such an architecture is susceptible to the information bottleneck issue. Some modifications have been made to the baseline encoder-decoder structure by introducing short-cuts~\cite{ronneberger2015u}; nevertheless, information exchange across different scales remains not very flexible.
Besides the global network architecture, the local information aggregation mechanism also plays an important role in facilitating information flows. For instance, trainable fusion rules that take into account the relative importance of incoming data streams are typically more effective than fixed ones.

The proposed DSCAN is designed to enable efficient information exchange and aggregation by employing 1) dense connections among different scales and 2) CSA-RDB, in which we enhance the widely used RDB~\cite{iebmrdn01} with channel-wise and spatial-wise attentions to differentiate features based on their relevance. The overall design effectively alleviates the bottleneck problem of multi-scale structure and helps unleash its potential for accomplishing challenging image restoration tasks. Specifically, the proposed DSCAN consists of three scales with 10, 8, and 4 CSA-RDBs, respectively. In order to connect different scales,  upsampling and downsampling need to be performed. Instead of using traditional interpolation methods (e.g., bilinear and bicubic), we adopt convolutional layers for feature map scale adjustment with the convolutional settings of stride and padding  chosen  in accordance with the needed downsampling/upsampling ratio. To effectively fuse the features produced from different scales, we capitalize on the Squeeze-and-Excitation (SE) module~\cite{hu2018squeeze} to perform feature addition with channel-wise attention.  Note that the SE module is light-weighted and only accounts for a negligible portion of the overall model size. Following~\cite{zhang2018density}, a global short-cut is introduced to direct the network to focus on rain/haze removal rather than reproducing texture details existent in the input image.

\begin{figure}[t]
	\centering
	\includegraphics[width=\linewidth]{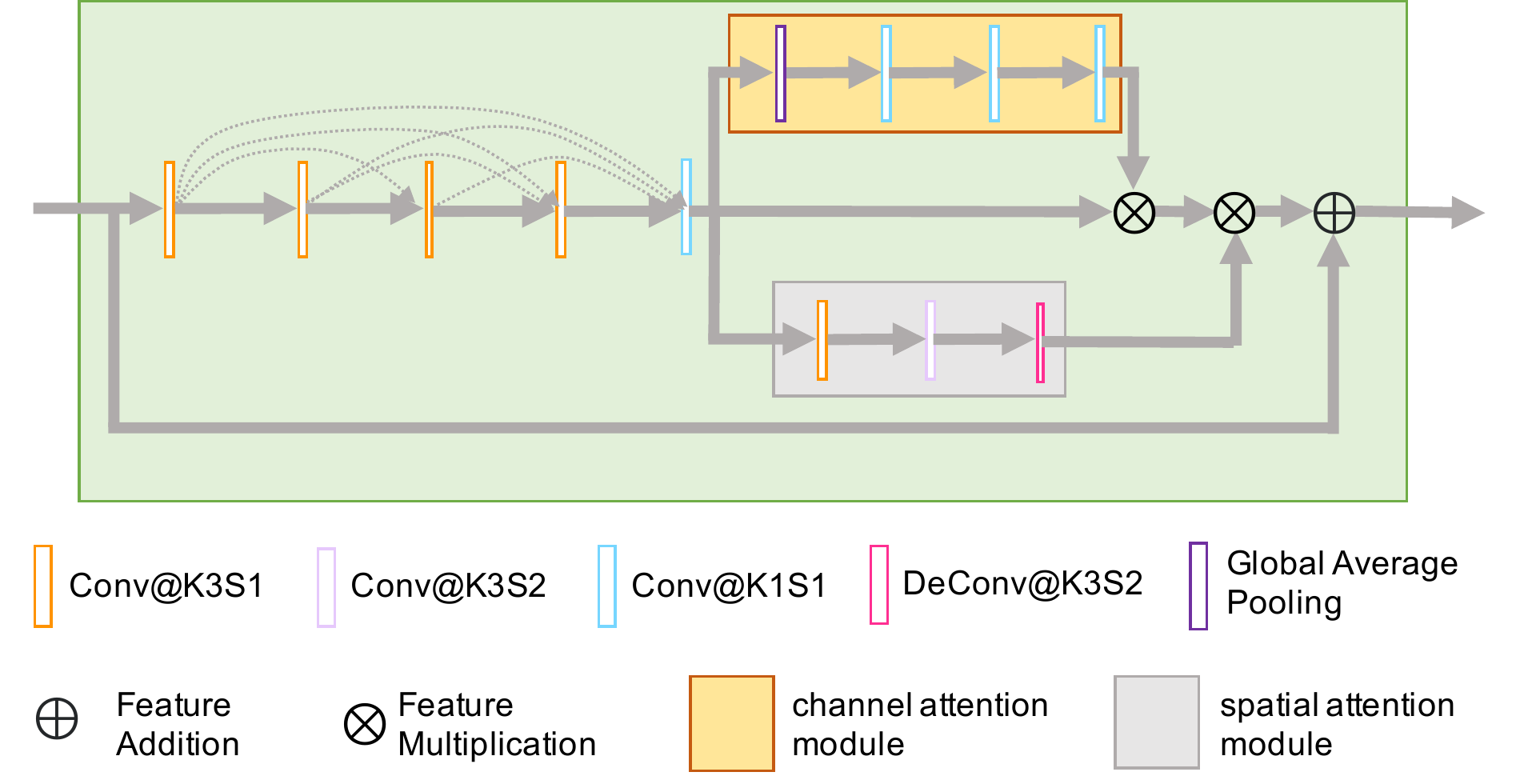}
	\caption{The detailed structure of CSA-RDB, where Conv(DeConv)@K$n$S$m$ denotes a $n \times n$ convolution (deconvolution) with stride $m$.}
	\label{fig:CSA-RDB}
	\figvspaceloose
\end{figure}

The detailed structure of CSA-RDB is shown in Fig.~\ref{fig:CSA-RDB}. We enhance the original version of RDB by introducing  channel-wise and spatial-wise attentions. The modified block can more effectively exploit different feature maps by leveraging the knowledge of  their relative importance learned from the network output via back propagation. The channel attention module in CSA-RDB is structurally the same as the SE module. The spatial attention module  produces a one-channel map assigning a different weight to each position  (the points at the same position of different channels share the same weight). The rest part of CSA-RDB follows the settings in~\cite{iebmrdn01}.



In DSCAN, all convolutional layers are activated by ReLU~\cite{nair2010rectified} except for the $1 \times 1$ convolutional layer (used to adjust the number of channels) in each CSA-RDB, and batch normalization is not used. We adopt the smooth $L_1$ loss in consideration of its advantage over the commonly used MSE loss in terms of the robustness to outliers and the ability to prevent gradient explosions in certain extreme cases~\cite{Girshick_2015_ICCV}. Let $\hat{B}^{(c)}(x)$ represent the intensity of pixel $x$ in color channel $c$ of derained/dehazed image $\hat{B}$, and let $N$ represent the total number of pixels in $\hat{B}$. The smooth $L_1$ loss is defined as
\begin{equation}
	L_{s} = \frac{1}{3N}\sum\limits_{c}\sum_x F_{s}(\hat{B}^{(c)}(x)-B^{(c)}(x)),
\end{equation}
where 
\begin{eqnarray}F_{s}(e)=
	\begin{cases}
		0.5e^2, &\mbox{if }|e|<1,\cr |e|-0.5, &\mbox{otherwise}. \cr\end{cases}
\end{eqnarray}


The proposed DSCAN is  model-agnostic in the sense that it does not attempt to explicitly estimate certain parameters in the rain model. We choose such design for the following reasons: 1) A model-dependent
deraining method  may have poor
generalization performance when applied to real-world images
due to model mismatch. 2) More importantly, it is
not necessarily advantageous to adopt model-dependent deraining
methods even if one only deals with synthetic images
generated by that model. This is because using model
constraints in network design may result in an undesirable
loss surface as shown in \cite{liu2019griddehazenet}.  On the other hand, our rain model and deraining method are not disconnected. After all, the proposed DSCAN is trained using synthetic images generated by the rain model and consequently its weights must carry, albeit in an implicit manner, some information regarding the model. It turns out that this type of indirect model exploration is more robust and effective. 
Indeed,
our experimental results indicate that the proposed model-agnostic
method outperforms the existing model-dependent
methods \cite{iebmmscnn01,wang2019spatial} on both synthetic and real-world images.

\section{Experimental Results}
Extensive experiments are conducted to demonstrate the competitive performance of the proposed DSCAN and validate the unified physical model for rain and haze effects. We also perform ablation studies to justify the overall design of DSCAN. 
Additional experimental results can be found in the supplementary material. The source code and our rainy dataset will be made publicly available. 


\subsection{Training and Testing Dataset}
Since collecting a large number of real rainy/hazy images and their clear counterpart is a formidable task, the data-driven deraining and dehazing methods often relies on synthetic images for training. Our unified physical model can be used to generate realistic-looking synthetic  rainy/hazy images based on clear images and their depth maps. For the dehazing problem, a large-scale dataset, named  RESIDE~\cite{li2019benchmarking}, was built based on ASM. Since our model degenerates to ASM when only the haze effect is concerned, we adopt RESIDE for synthetic hazy data to avoid unnecessary duplication.

For the deraining problem, a benchmark was built in~\cite{li2019single} using an over-simplified model with a single-layer rain streak superimposed on the clear image (i.e., (\ref{equ:basic})). In this work we use our physical model to construct a new synthetic dataset of rainy images as follows. We carefully select 1000 clear images (accompanied by their depth maps) from the Outdoor Training Set (OTS)~\cite{li2019benchmarking}. Each clear image is utilized to generate 14 rainy versions. In total, we generate 12600 rainy images (from 900 clear images) for training and 1400 rainy images (from 100 clear images) for testing. To simulate different rain densities, the normalized mean $\mu$ of Poisson point process is randomly chosen from $\left[0.005, 0.05\right]$. For the rain kernel, we sample the streak length $l_{s}$ from $ \left[0.05 s, 0.2 s\right]$, the streak width $w_{s}$ from $\left[0.005 s, 0.025 s \right]$, and the streak direction $d_{s}$ from $\left[-30^{\circ}, 30^{\circ}\right]$, where $s$ is the minimum of image height and width. We choose the global atmospheric light $A \in \left[0.7, 1.0 \right]$ and the attenuation ratio $\alpha \in \left[0.6, 0.9\right]$. In addition to synthetic images generated by our physical model, we also use real rainy and hazy images from~\cite{li2019single} and~\cite{li2019benchmarking}  for qualitative comparisons.

\begin{figure}[!t]
	\centering
	\begin{minipage}[h]{0.49\linewidth}
		\centering
		\includegraphics[width=\linewidth]{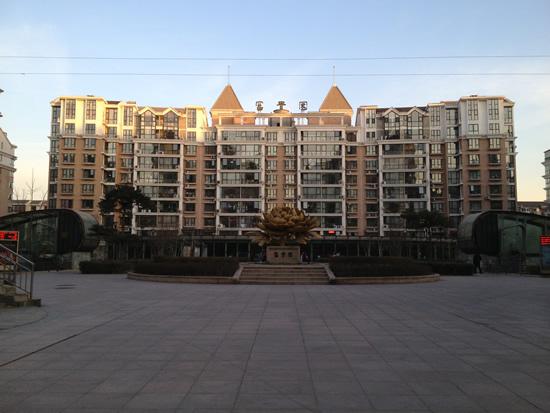}
		\scriptsize{(a) Clear image}
	\end{minipage}	
	\begin{minipage}[h]{0.49\linewidth}
		\centering
		\includegraphics[width=\linewidth]{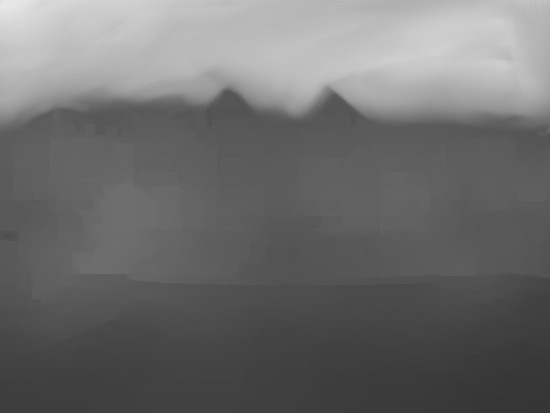}
		\scriptsize{(b) Depth map}
	\end{minipage}
	\begin{minipage}[h]{0.49\linewidth}
		\centering
		\includegraphics[width=\linewidth]{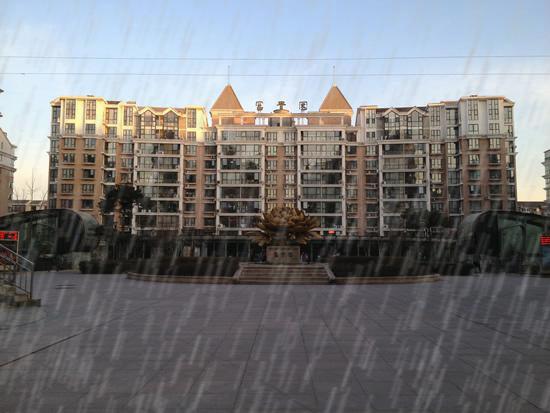}
		\scriptsize{(c) Benchmark~\cite{li2019single}}
	\end{minipage}	
	\begin{minipage}[h]{0.49\linewidth}
		\centering
		\includegraphics[width=\linewidth]{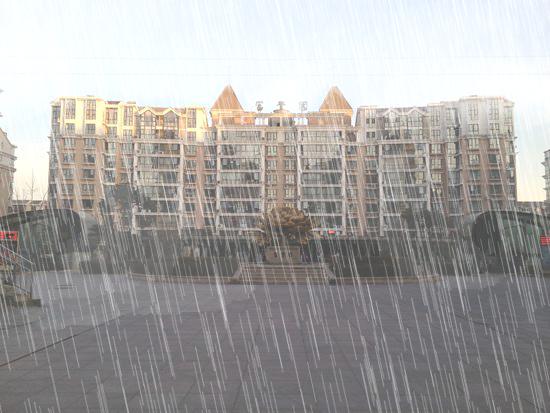}
		\scriptsize{(d) Ours}
	\end{minipage}	
	
	\caption{Comparison of a synthetic rainy image in benchmark~\cite{li2019single} based on the over-simplified model (\ref{equ:basic}) and a version generated using our new physical model.}
	\label{fig:rainmodels}
\end{figure}

In Fig.~\ref{fig:rainmodels}, we compare a synthetic rainy image in benchmark~\cite{li2019single} (which is generated using the over-simplified model (\ref{equ:basic})) with a version generated using our new physical model. It can be seen that our rainy image looks more realistic, which can be attributed to depth-aware fusion of multi-layer rain streaks accommodated by the proposed physical model. In particular, the haze effect naturally emerges in distant areas of our synthetic image  even though it is not explicitly built into the proposed model.

\subsection{Implementation} \label{implementation}
We train the proposed DSCAN end-to-end with cropped sRGB image patches of size $240 \times 240$, accelerated by optimizer~\cite{kingma2014adam} (batch size $=10$, $\beta_1=0.9$ and $\beta_2=0.999$). The network is trained for 50 epochs in total. We set the initial learning rate to 0.001 and reduce the learning rate by half every 10 epochs.  We conduct  training and testing on a PC with two NVIDIA GeForce GTX 1080Ti, and adopt PSNR and SSIM as metrics for objective assessment.


\begin{figure*}[t]
	\centering
	\begin{minipage}[h]{0.105\linewidth}
		\centering
		\includegraphics[width=\linewidth]{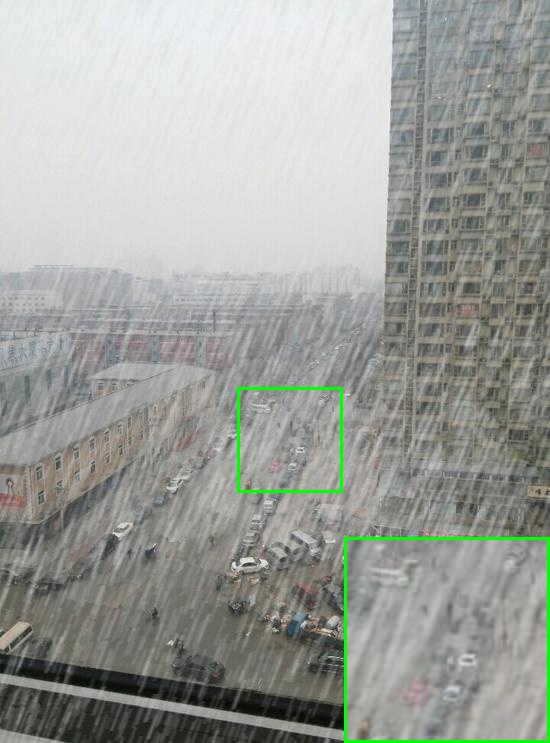}
	\end{minipage}
	\begin{minipage}[h]{0.105\linewidth}
		\centering
		\includegraphics[width=\linewidth]{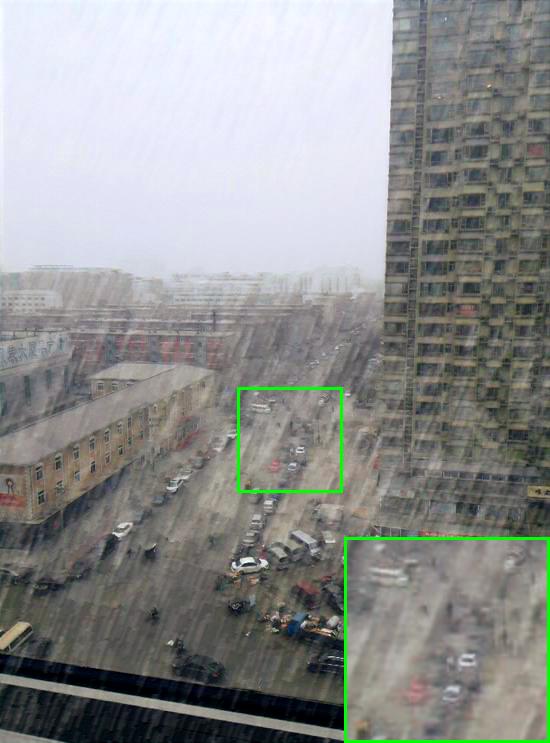}
	\end{minipage}
	\begin{minipage}[h]{0.105\linewidth}
		\centering
		\includegraphics[width=\linewidth]{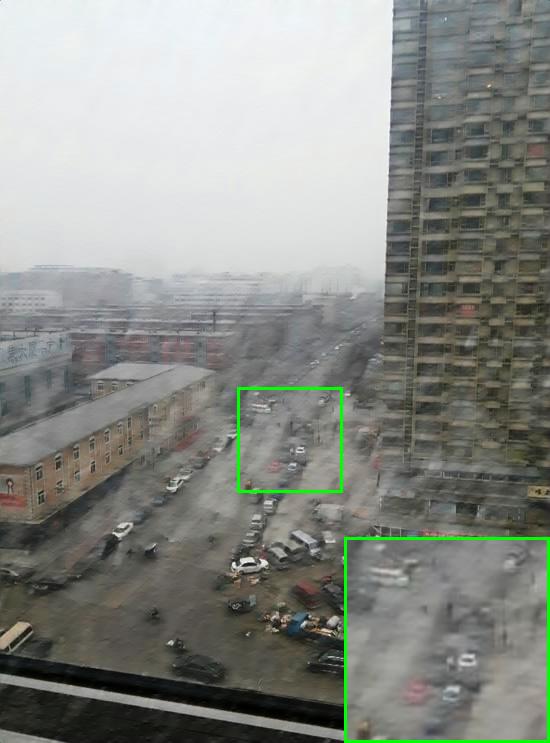}
	\end{minipage}
	\begin{minipage}[h]{0.105\linewidth}
		\centering
		\includegraphics[width=\linewidth]{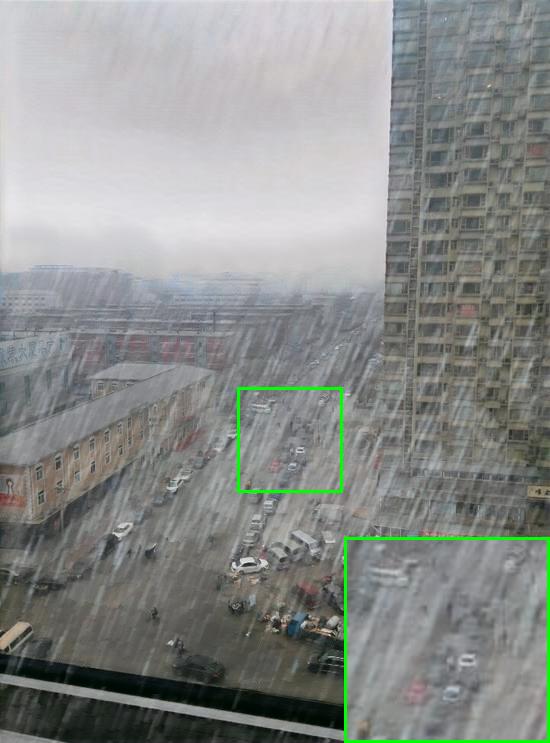}
	\end{minipage}		
	\begin{minipage}[h]{0.105\linewidth}
		\centering
		\includegraphics[width=\linewidth]{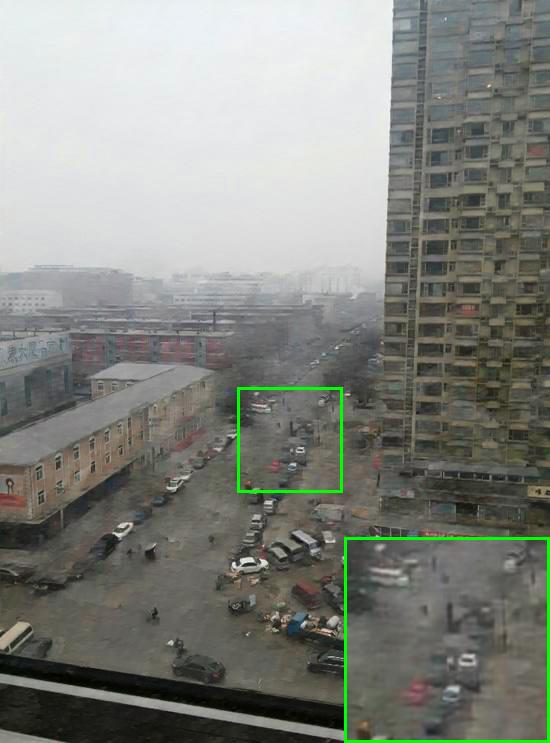}
	\end{minipage}		
	\begin{minipage}[h]{0.105\linewidth}
		\centering
		\includegraphics[width=\linewidth]{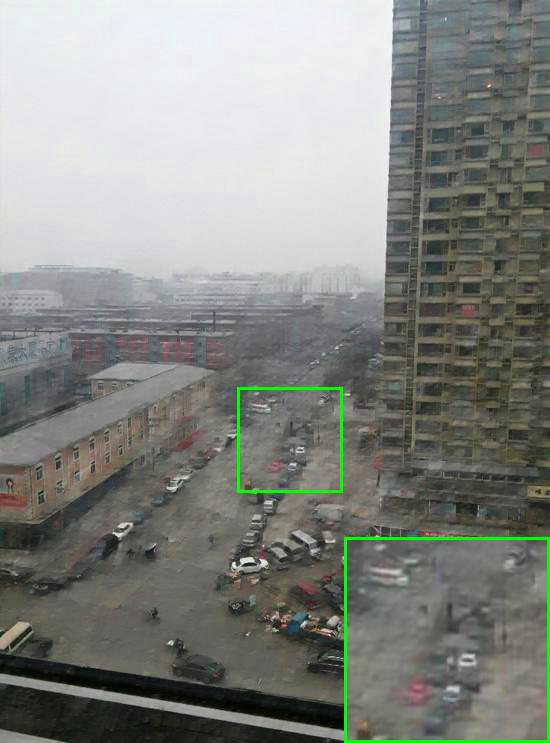}
	\end{minipage}	
	\begin{minipage}[h]{0.105\linewidth}
		\centering
		\includegraphics[width=\linewidth]{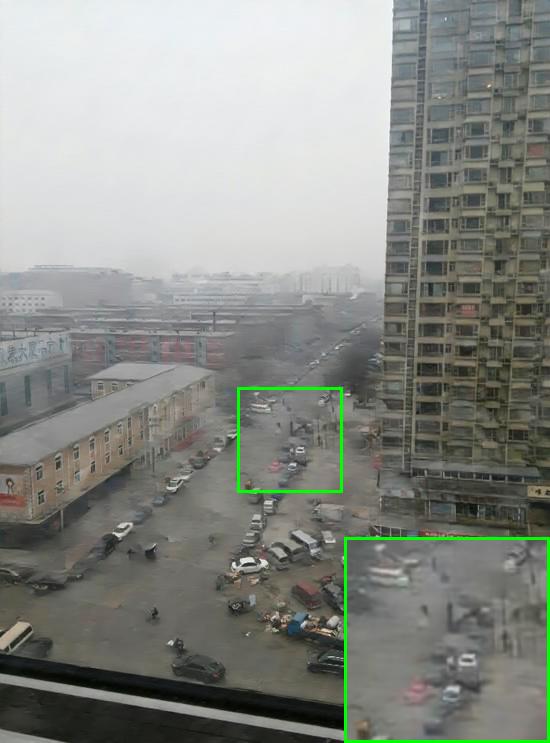}
	\end{minipage}	
	\begin{minipage}[h]{0.105\linewidth}
		\centering
		\includegraphics[width=\linewidth]{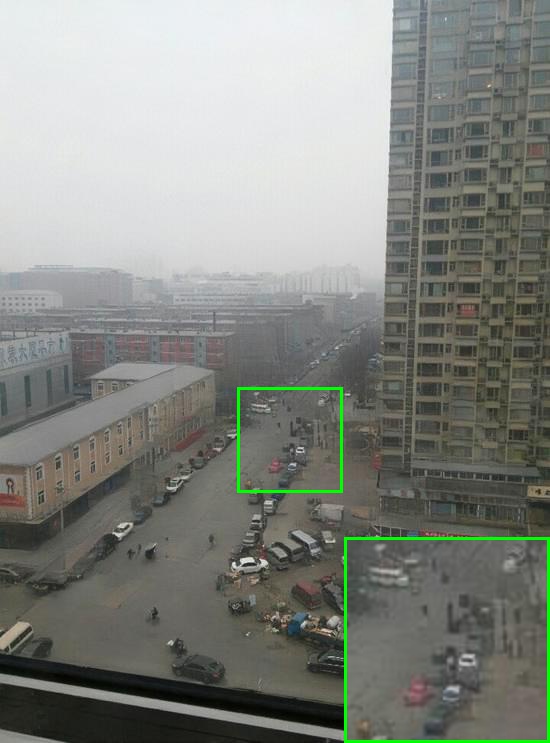}
	\end{minipage}	
	\begin{minipage}[h]{0.105\linewidth}
		\centering
		\includegraphics[width=\linewidth]{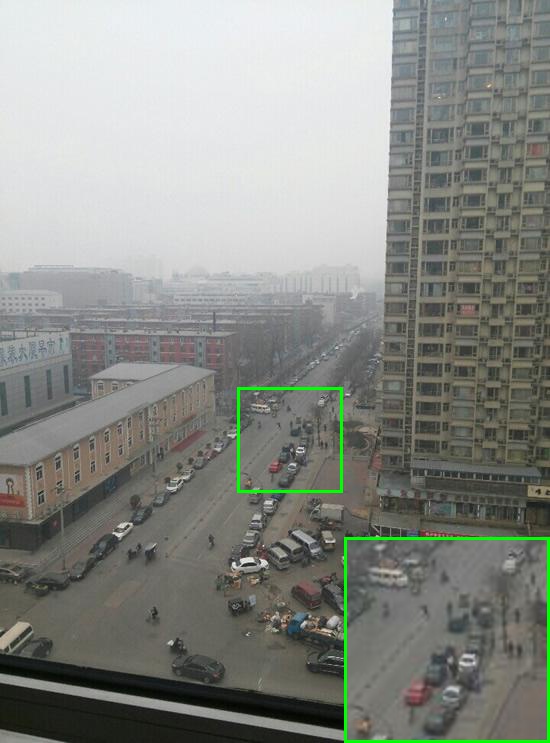}
	\end{minipage}
	\begin{minipage}[h]{0.105\linewidth}
		\centering
		\includegraphics[width=\linewidth]{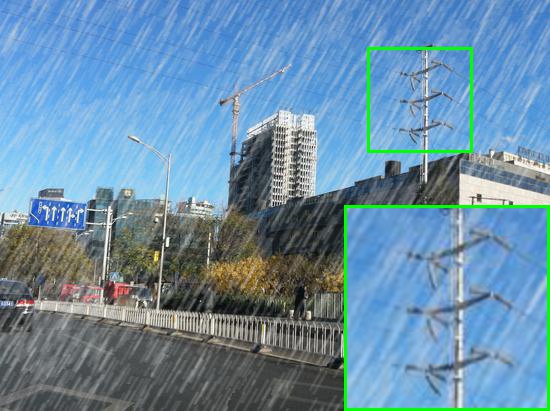}
	\end{minipage}
	\begin{minipage}[h]{0.105\linewidth}
		\centering
		\includegraphics[width=\linewidth]{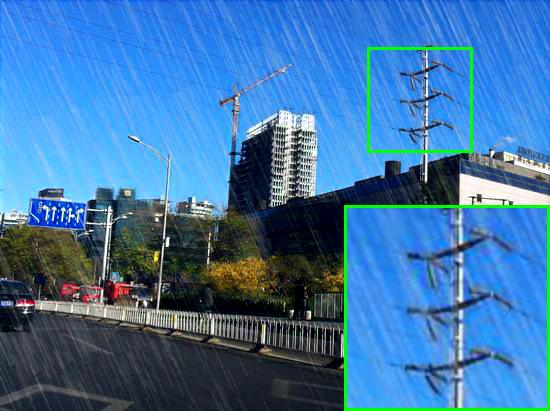}
	\end{minipage}
	\begin{minipage}[h]{0.105\linewidth}
		\centering
		\includegraphics[width=\linewidth]{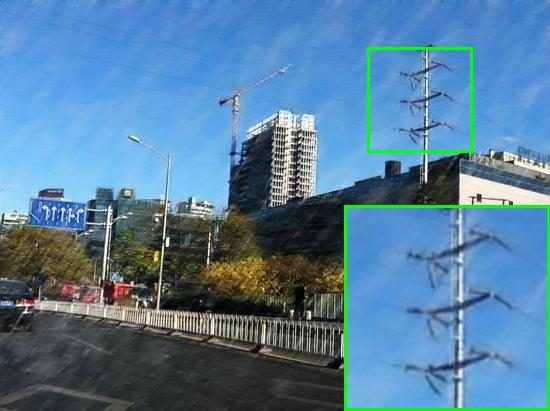}
	\end{minipage}
	\begin{minipage}[h]{0.105\linewidth}
		\centering
		\includegraphics[width=\linewidth]{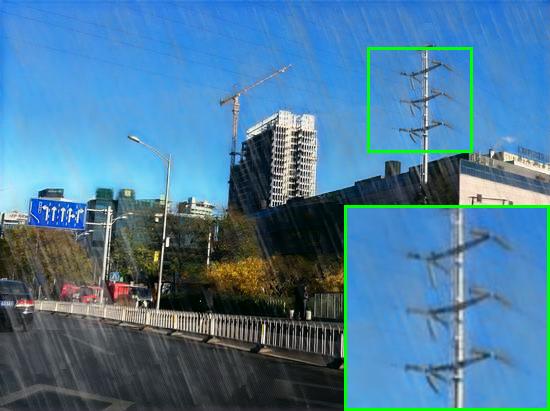}
	\end{minipage}		
	\begin{minipage}[h]{0.105\linewidth}
		\centering
		\includegraphics[width=\linewidth]{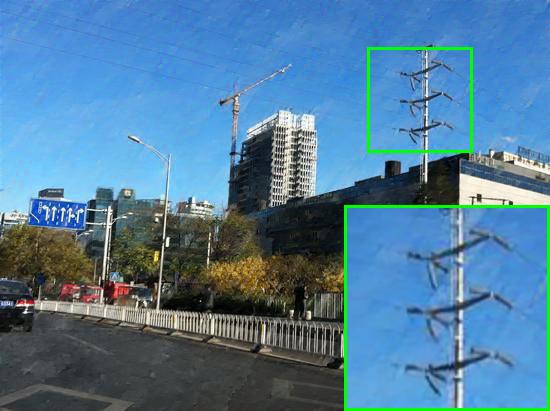}
	\end{minipage}		
	\begin{minipage}[h]{0.105\linewidth}
		\centering
		\includegraphics[width=\linewidth]{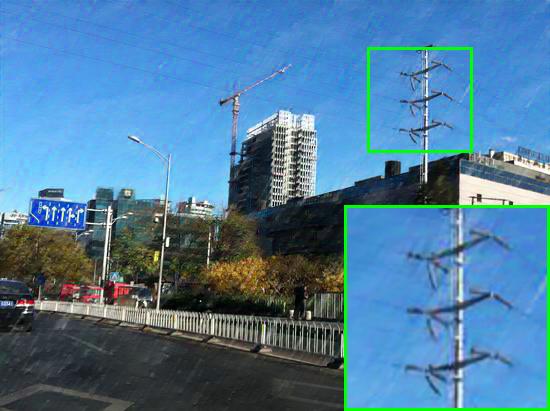}
	\end{minipage}	
	\begin{minipage}[h]{0.105\linewidth}
		\centering
		\includegraphics[width=\linewidth]{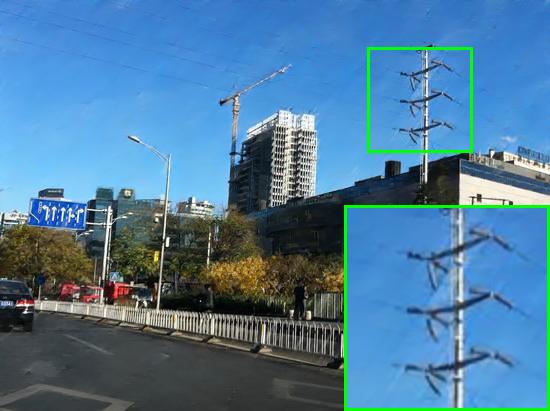}
	\end{minipage}	
	\begin{minipage}[h]{0.105\linewidth}
		\centering
		\includegraphics[width=\linewidth]{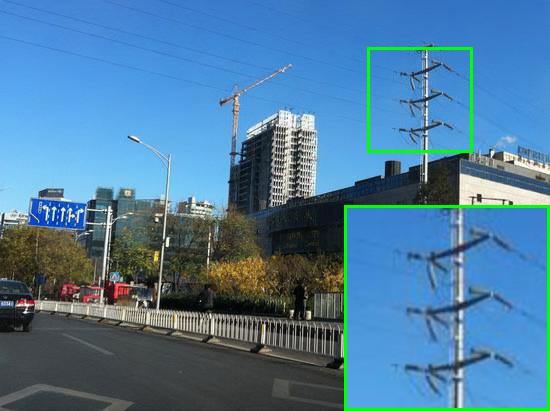}
	\end{minipage}	
	\begin{minipage}[h]{0.105\linewidth}
		\centering
		\includegraphics[width=\linewidth]{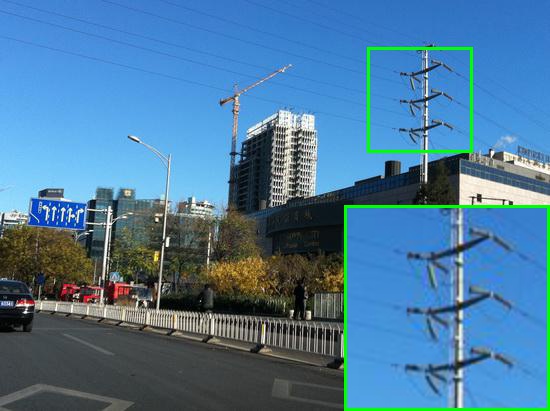}
	\end{minipage}	
	\begin{minipage}[h]{0.105\linewidth}
		\centering
		\includegraphics[width=\linewidth]{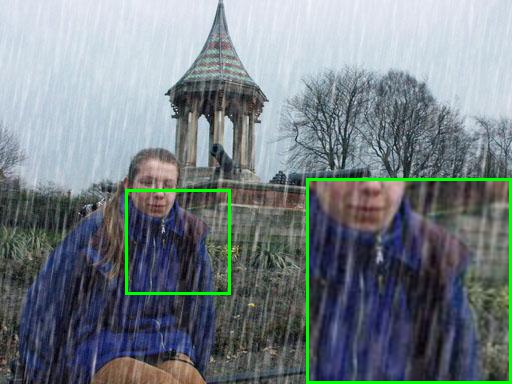}
	\end{minipage}
	\begin{minipage}[h]{0.105\linewidth}
		\centering
		\includegraphics[width=\linewidth]{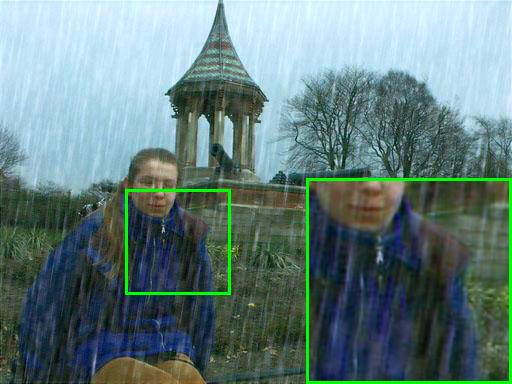}
	\end{minipage}
	\begin{minipage}[h]{0.105\linewidth}
		\centering
		\includegraphics[width=\linewidth]{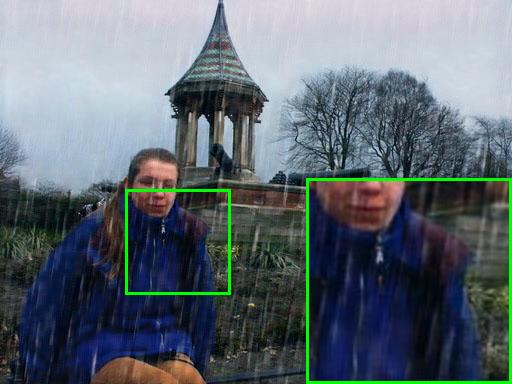}
	\end{minipage}	
	\begin{minipage}[h]{0.105\linewidth}
		\centering
		\includegraphics[width=\linewidth]{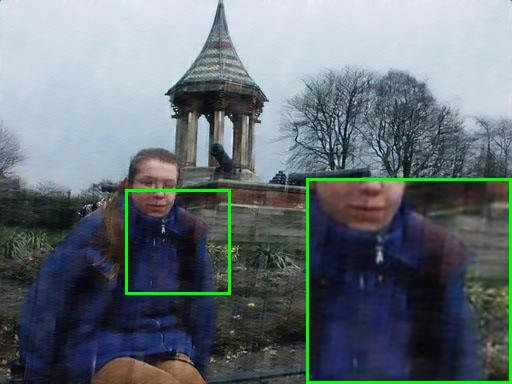}
	\end{minipage}	
	\begin{minipage}[h]{0.105\linewidth}
		\centering
		\includegraphics[width=\linewidth]{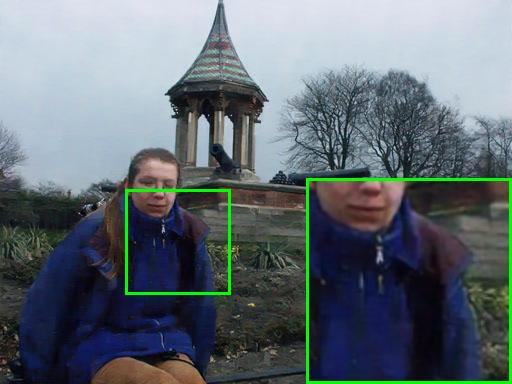}
	\end{minipage}		
	\begin{minipage}[h]{0.105\linewidth}
		\centering
		\includegraphics[width=\linewidth]{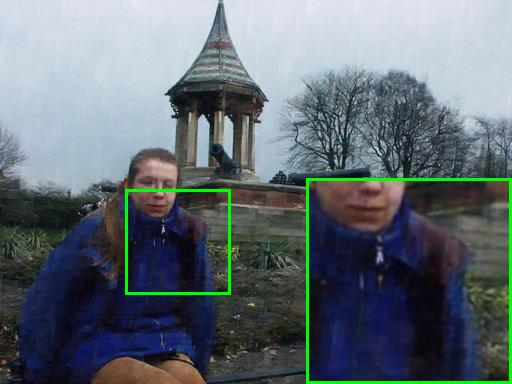}
	\end{minipage}	
	\begin{minipage}[h]{0.105\linewidth}
		\centering
		\includegraphics[width=\linewidth]{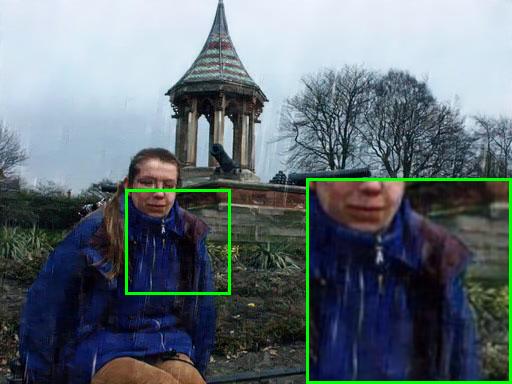}
	\end{minipage}	
	\begin{minipage}[h]{0.105\linewidth}
		\centering
		\includegraphics[width=\linewidth]{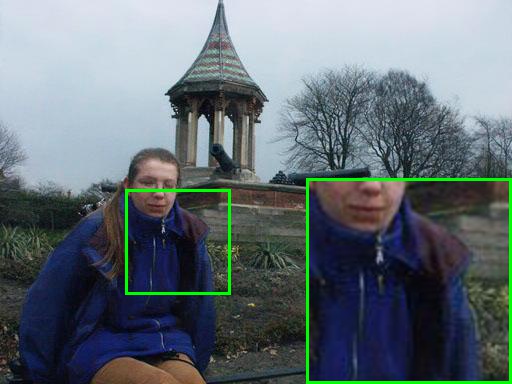}
	\end{minipage}	
	\begin{minipage}[h]{0.105\linewidth}
		\centering
		\includegraphics[width=\linewidth]{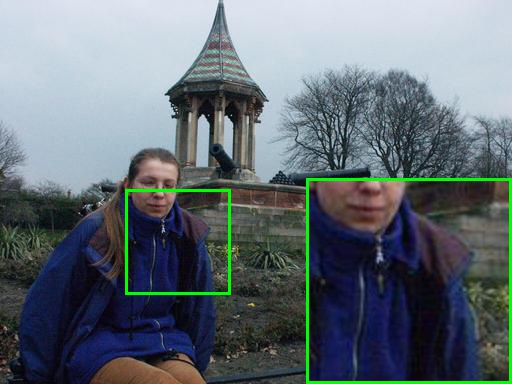}
	\end{minipage}				
	\begin{minipage}[h]{0.105\linewidth}
		\centering
		\includegraphics[width=\linewidth]{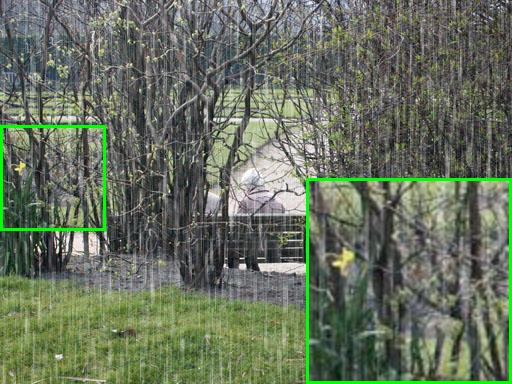}
		\scriptsize{(a) Rainy}
	\end{minipage}
	\begin{minipage}[h]{0.105\linewidth}
		\centering
		\includegraphics[width=\linewidth]{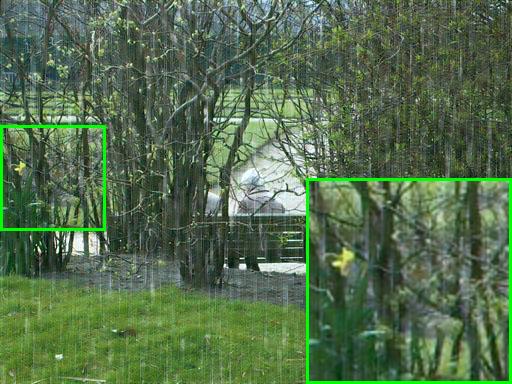}
		\scriptsize{(b) DDN}
	\end{minipage}
	\begin{minipage}[h]{0.105\linewidth}
		\centering
		\includegraphics[width=\linewidth]{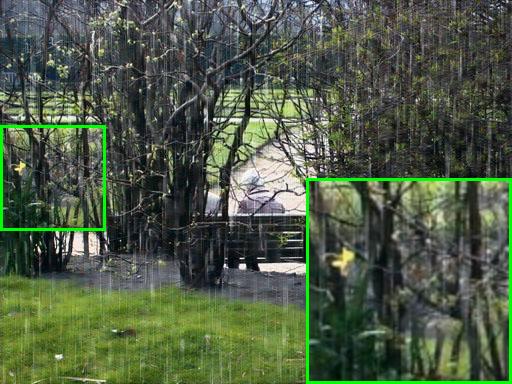}
		\scriptsize{(c) DID-MDN}
	\end{minipage}	
	\begin{minipage}[h]{0.105\linewidth}
		\centering
		\includegraphics[width=\linewidth]{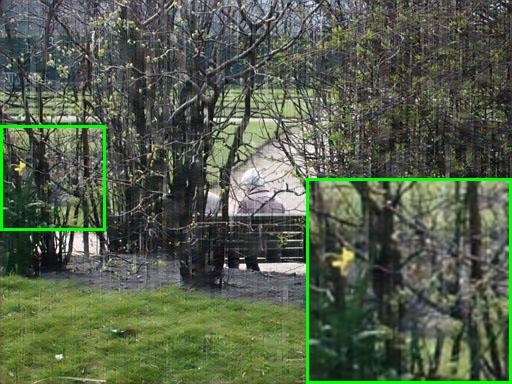}
		\scriptsize{(d) JORDER}
	\end{minipage}	
	\begin{minipage}[h]{0.105\linewidth}
		\centering
		\includegraphics[width=\linewidth]{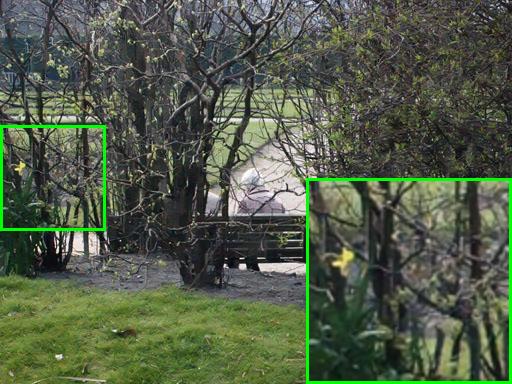}
		\scriptsize{(e) PReNet}
	\end{minipage}		
	\begin{minipage}[h]{0.105\linewidth}
		\centering
		\includegraphics[width=\linewidth]{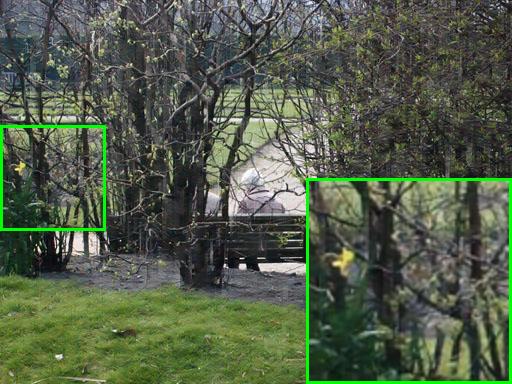}
		\scriptsize{(f) RESCAN}
	\end{minipage}	
	\begin{minipage}[h]{0.105\linewidth}
		\centering
		\includegraphics[width=\linewidth]{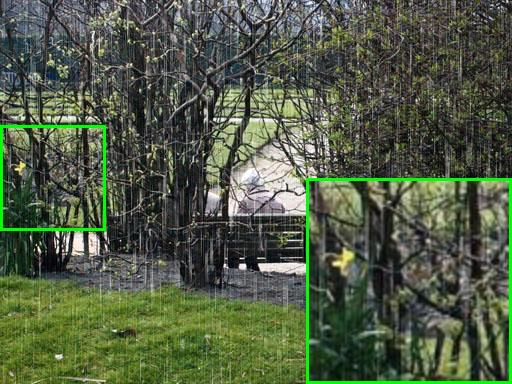}
		\scriptsize{(g) SPANet}
	\end{minipage}	
	\begin{minipage}[h]{0.105\linewidth}
		\centering
		\includegraphics[width=\linewidth]{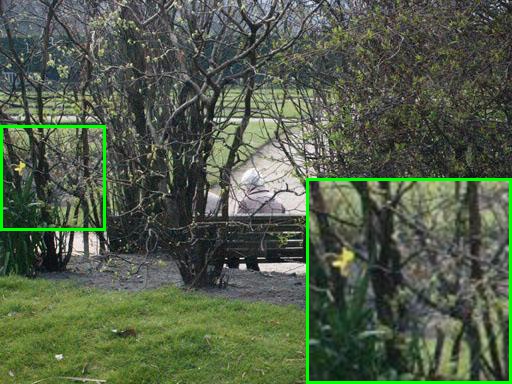}
		\scriptsize{(h) Ours}
	\end{minipage}		
	\begin{minipage}[h]{0.105\linewidth}
		\centering
		\includegraphics[width=\linewidth]{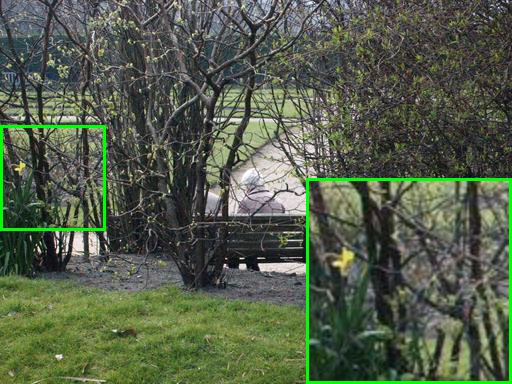}
		\scriptsize{(i) GT}
	\end{minipage}		
	\caption{Qualitative comparisons on synthetic rainy images generated by our unified physical model.}
	\label{fig:synthetic_rain}
	\figvspaceloose
\end{figure*}

\begin{figure*}[t]
	\centering
	\begin{minipage}[h]{0.117\linewidth}
		\centering
		\includegraphics[width=\linewidth]{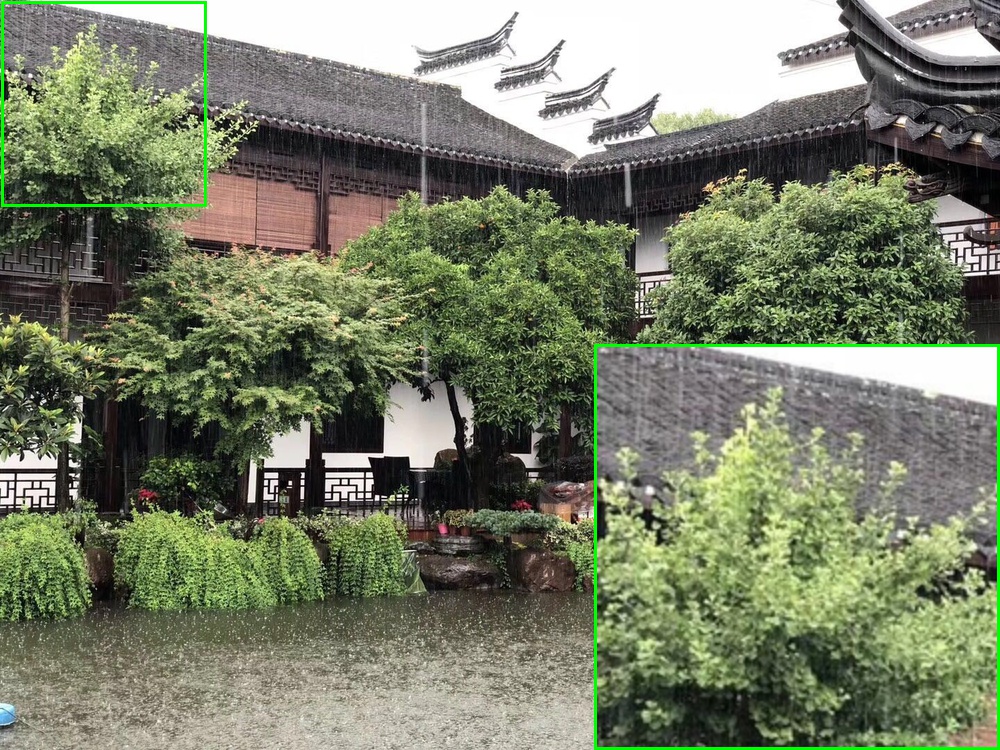}
	\end{minipage}
	\begin{minipage}[h]{0.117\linewidth}
		\centering
		\includegraphics[width=\linewidth]{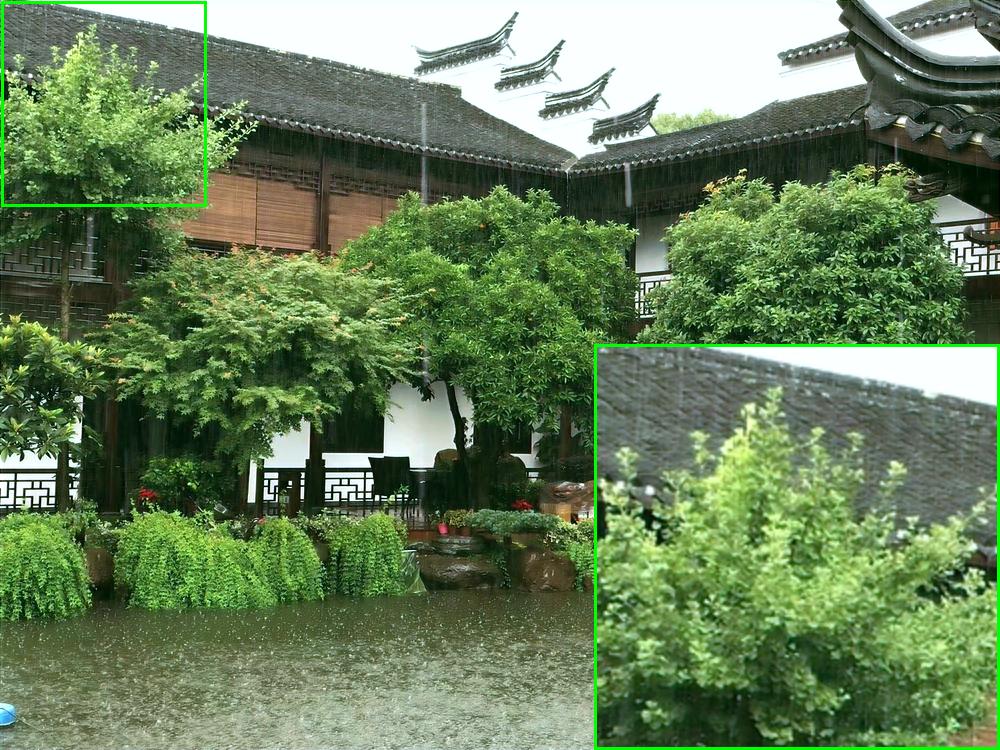}
	\end{minipage}
	\begin{minipage}[h]{0.117\linewidth}
		\centering
		\includegraphics[width=\linewidth]{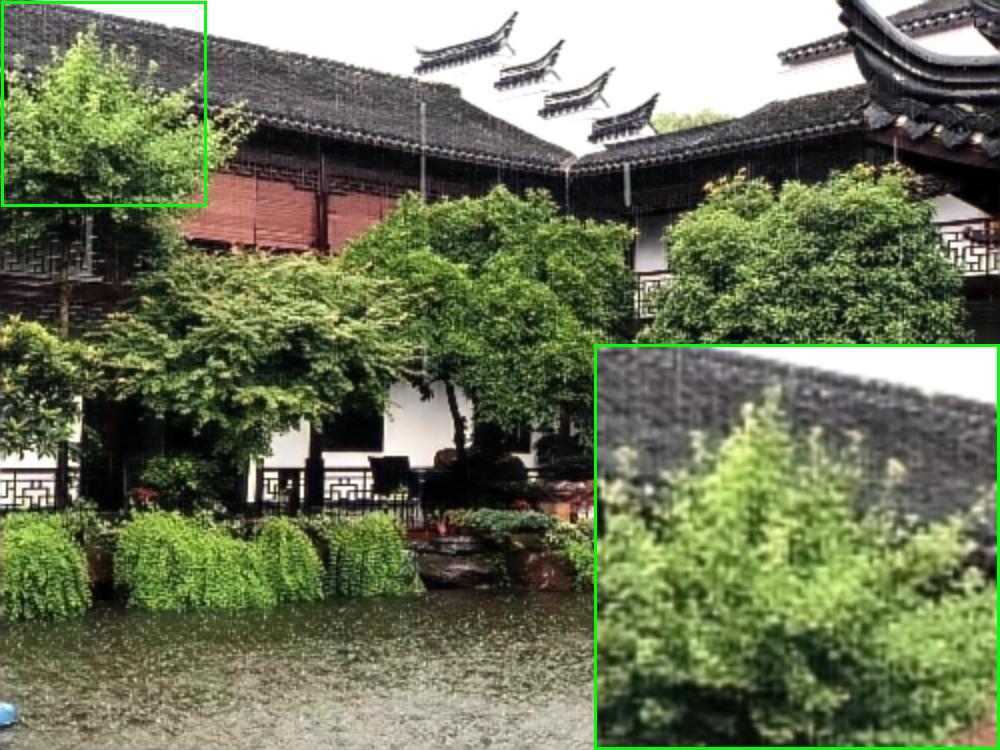}
	\end{minipage}
	\begin{minipage}[h]{0.117\linewidth}
		\centering
		\includegraphics[width=\linewidth]{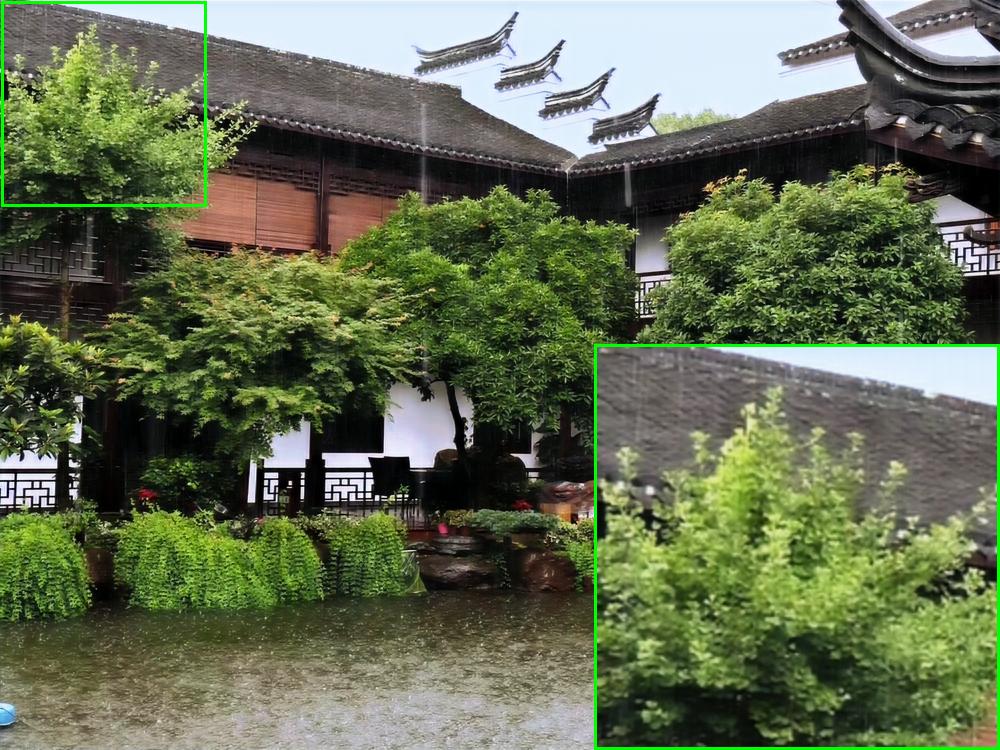}
	\end{minipage}	
	\begin{minipage}[h]{0.117\linewidth}
		\centering
		\includegraphics[width=\linewidth]{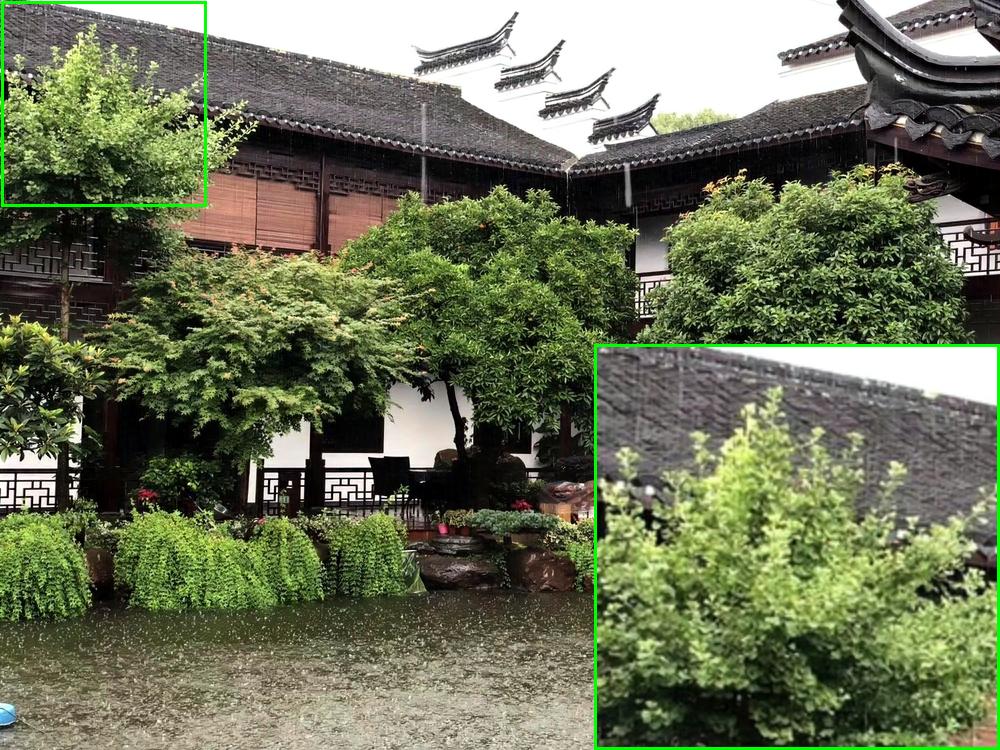}
	\end{minipage}		
	\begin{minipage}[h]{0.117\linewidth}
		\centering
		\includegraphics[width=\linewidth]{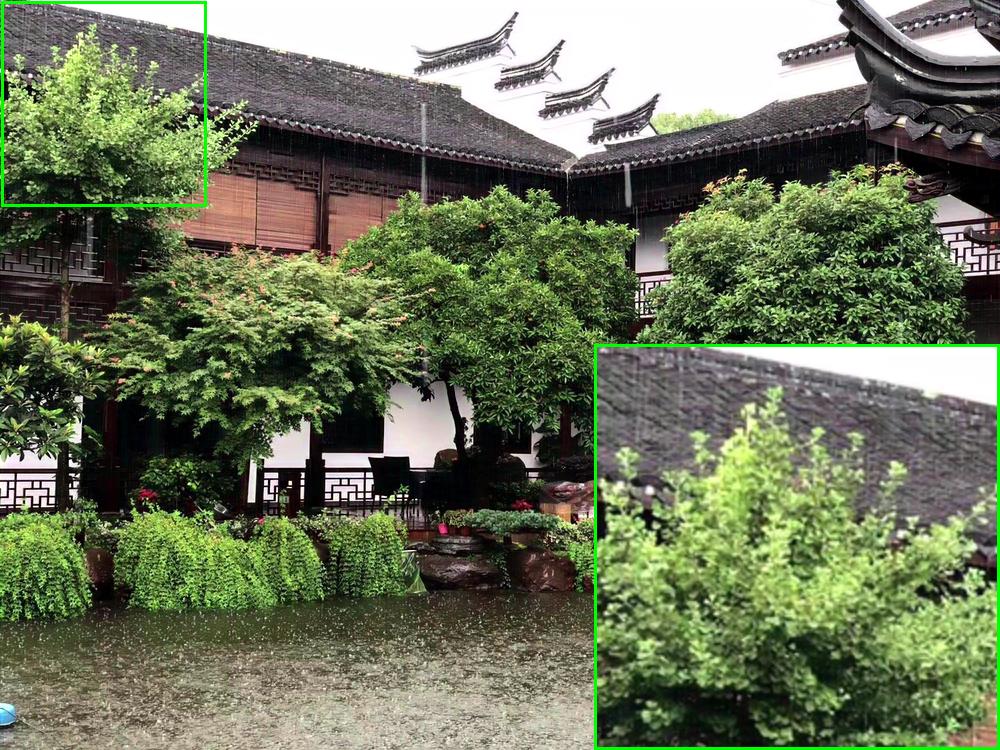}
	\end{minipage}	
	\begin{minipage}[h]{0.117\linewidth}
		\centering
		\includegraphics[width=\linewidth]{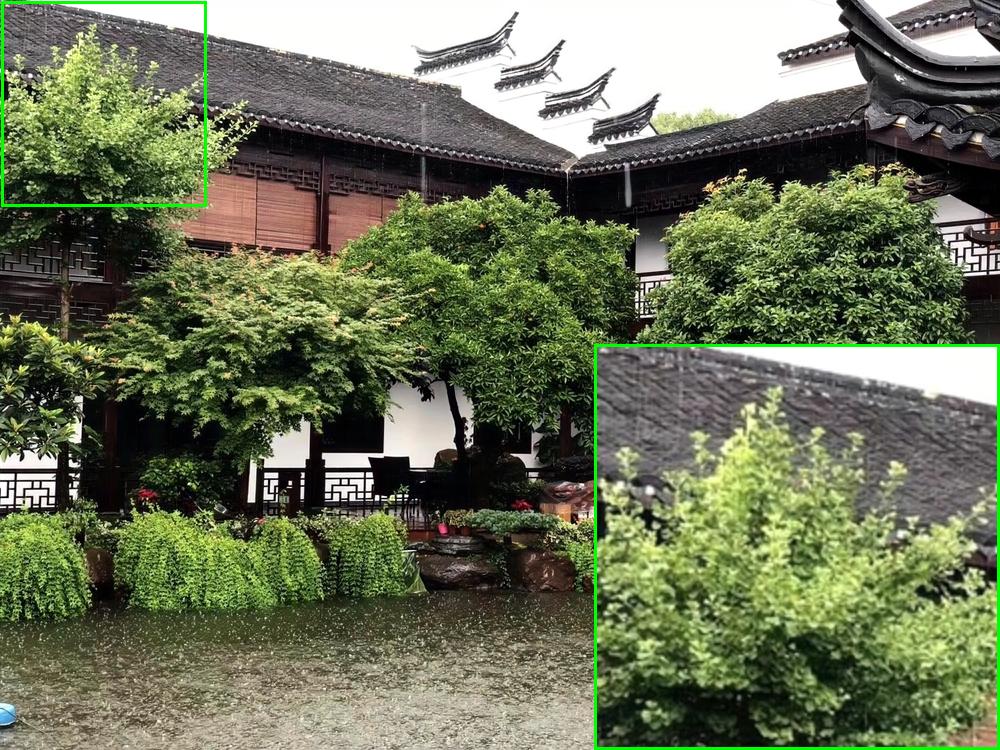}
	\end{minipage}	
	\begin{minipage}[h]{0.117\linewidth}
		\centering
		\includegraphics[width=\linewidth]{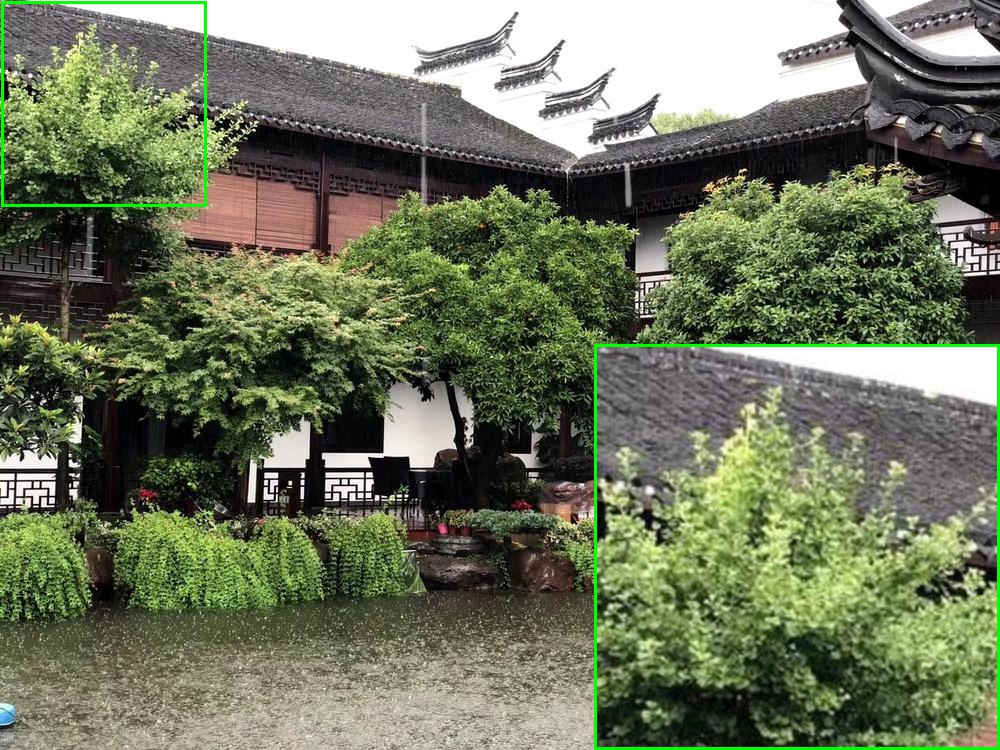}
	\end{minipage}		
	\begin{minipage}[h]{0.117\linewidth}
		\centering
		\includegraphics[width=\linewidth]{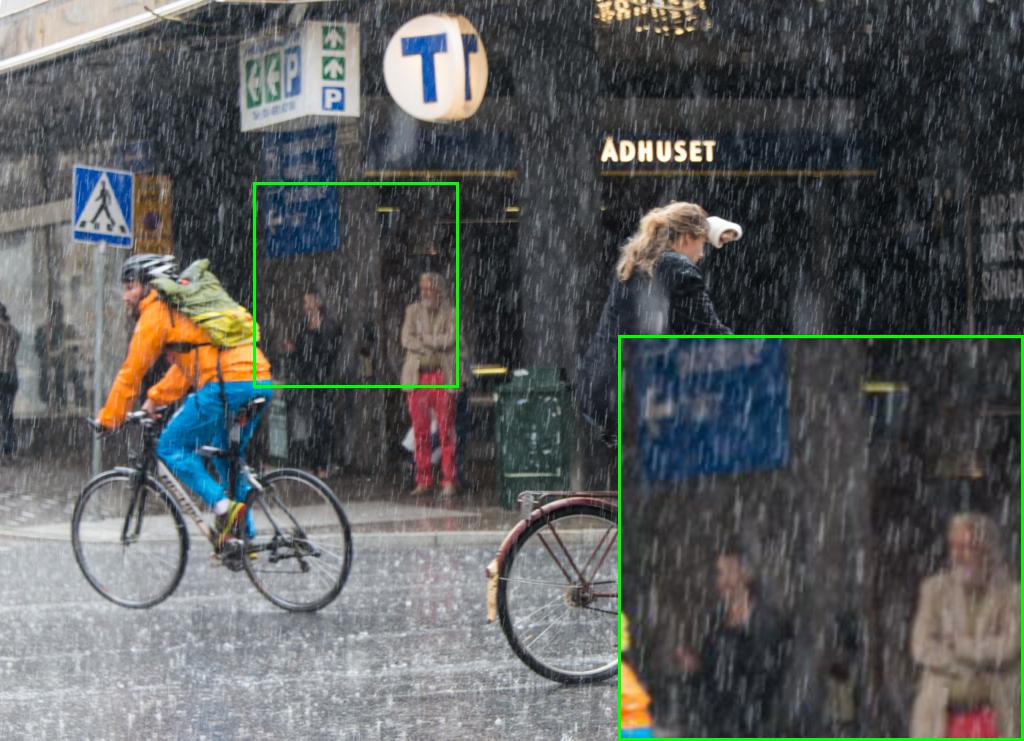}
		\scriptsize{(a) Rainy}
	\end{minipage}
	\begin{minipage}[h]{0.117\linewidth}
		\centering
		\includegraphics[width=\linewidth]{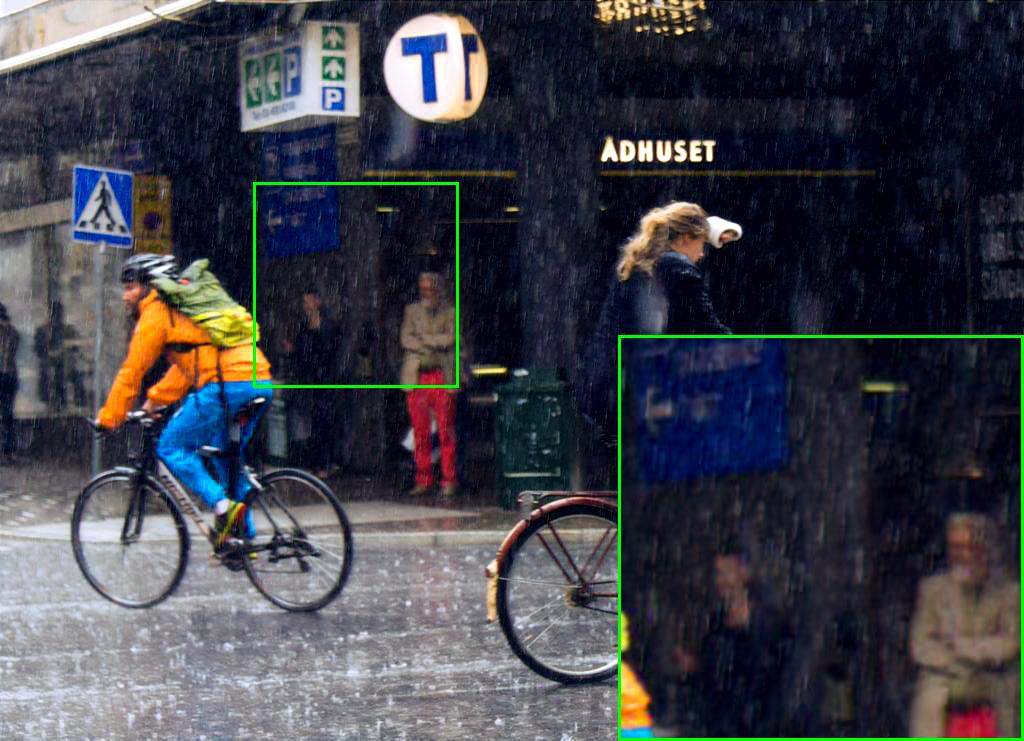}
		\scriptsize{(b) DDN}
	\end{minipage}
	\begin{minipage}[h]{0.117\linewidth}
		\centering
		\includegraphics[width=\linewidth]{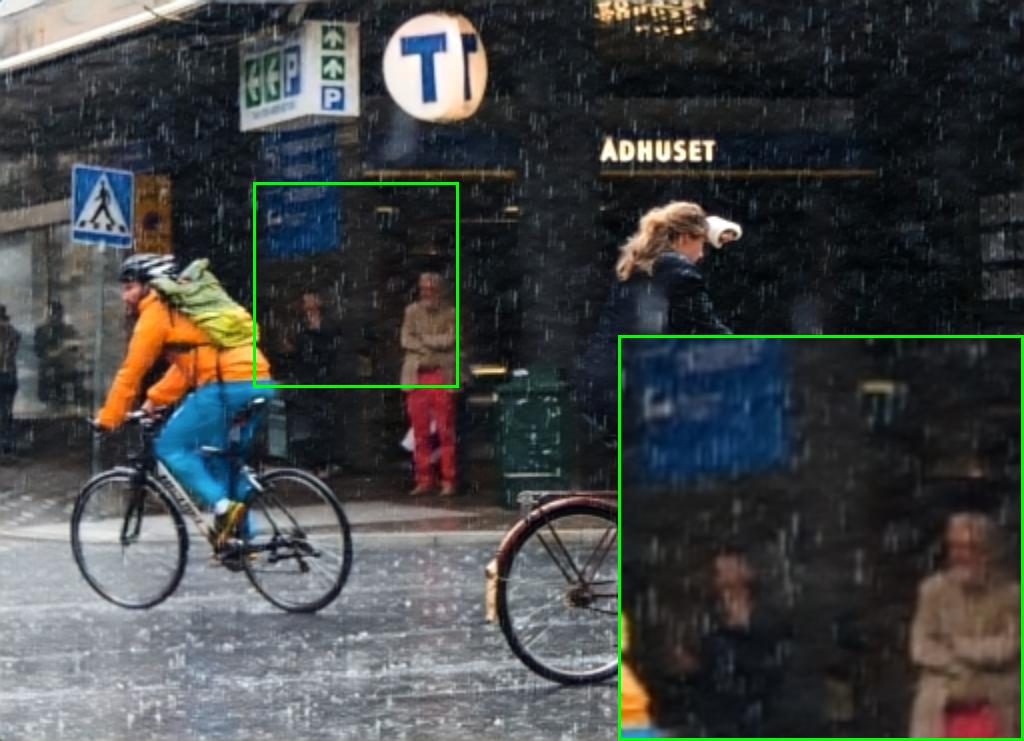}
		\scriptsize{(c) JORDER}
	\end{minipage}	
	\begin{minipage}[h]{0.117\linewidth}
		\centering
		\includegraphics[width=\linewidth]{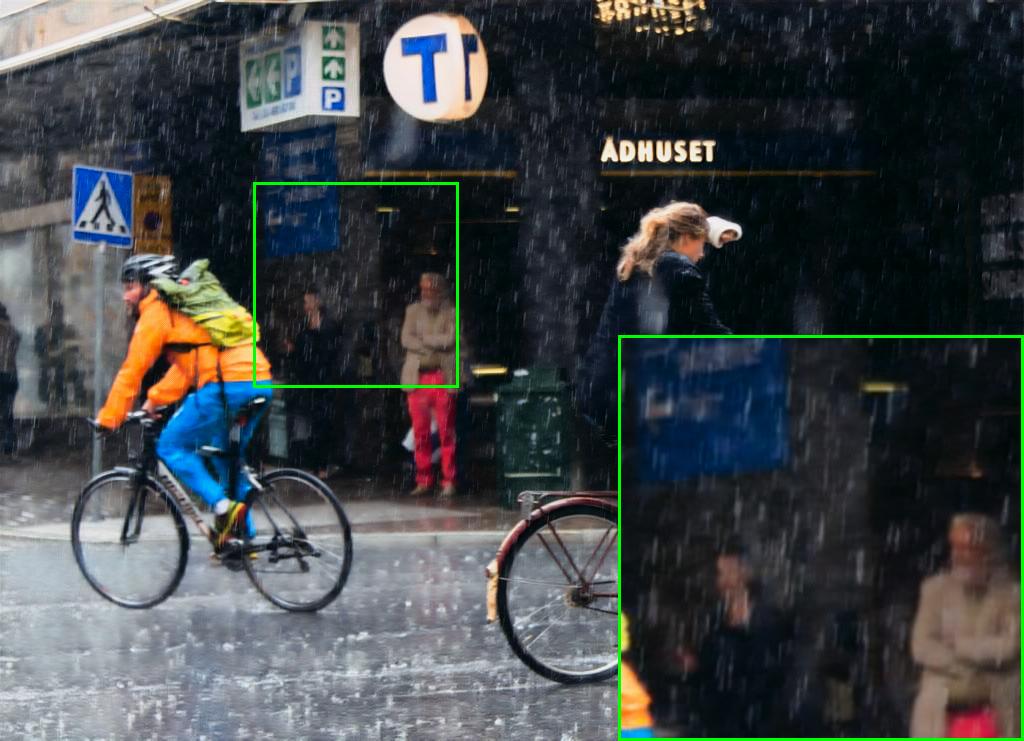}
		\scriptsize{(d) DID-MDN}
	\end{minipage}	
	\begin{minipage}[h]{0.117\linewidth}
		\centering
		\includegraphics[width=\linewidth]{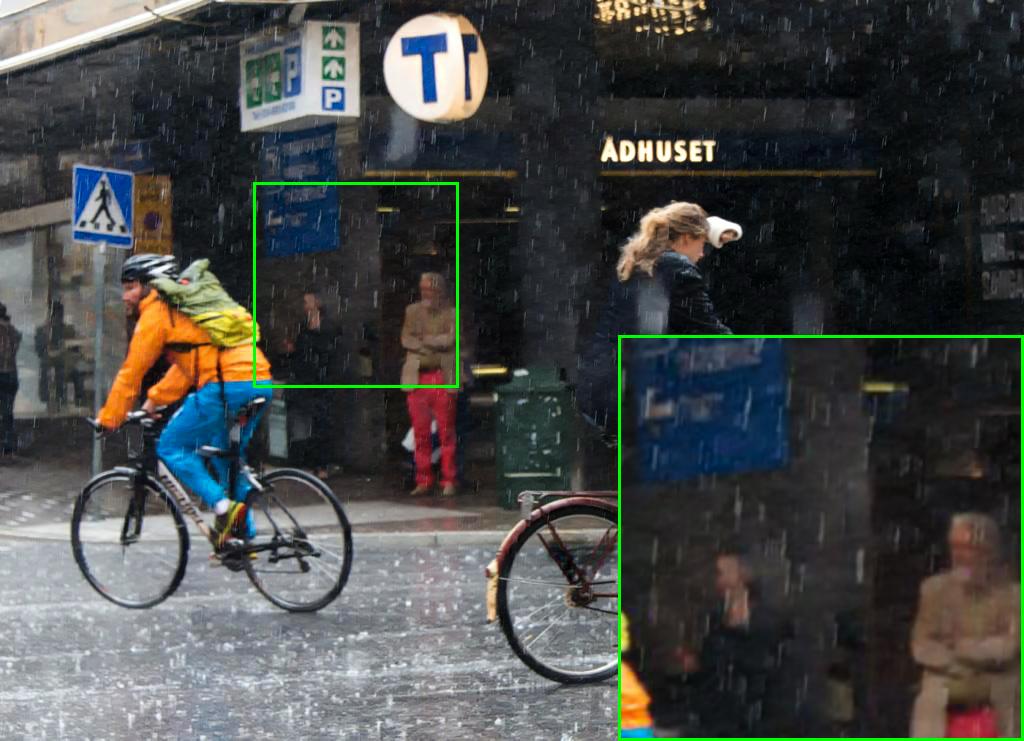}
		\scriptsize{(e) PReNet}
	\end{minipage}		
	\begin{minipage}[h]{0.117\linewidth}
		\centering
		\includegraphics[width=\linewidth]{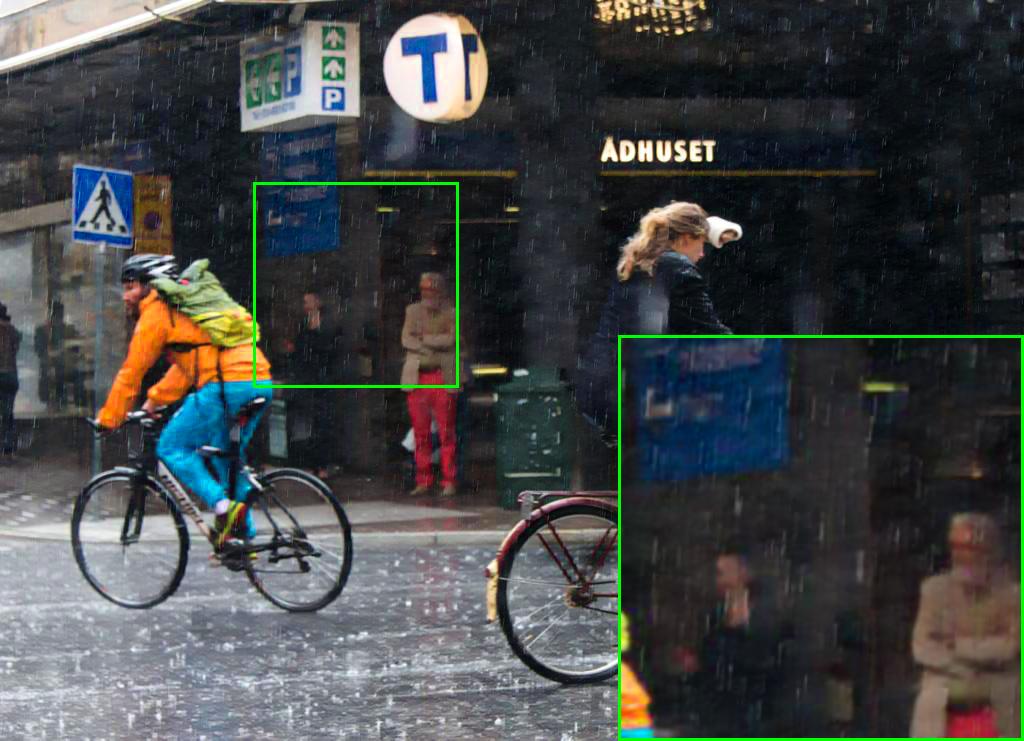}
		\scriptsize{(f) RESCAN}
	\end{minipage}	
	\begin{minipage}[h]{0.117\linewidth}
		\centering
		\includegraphics[width=\linewidth]{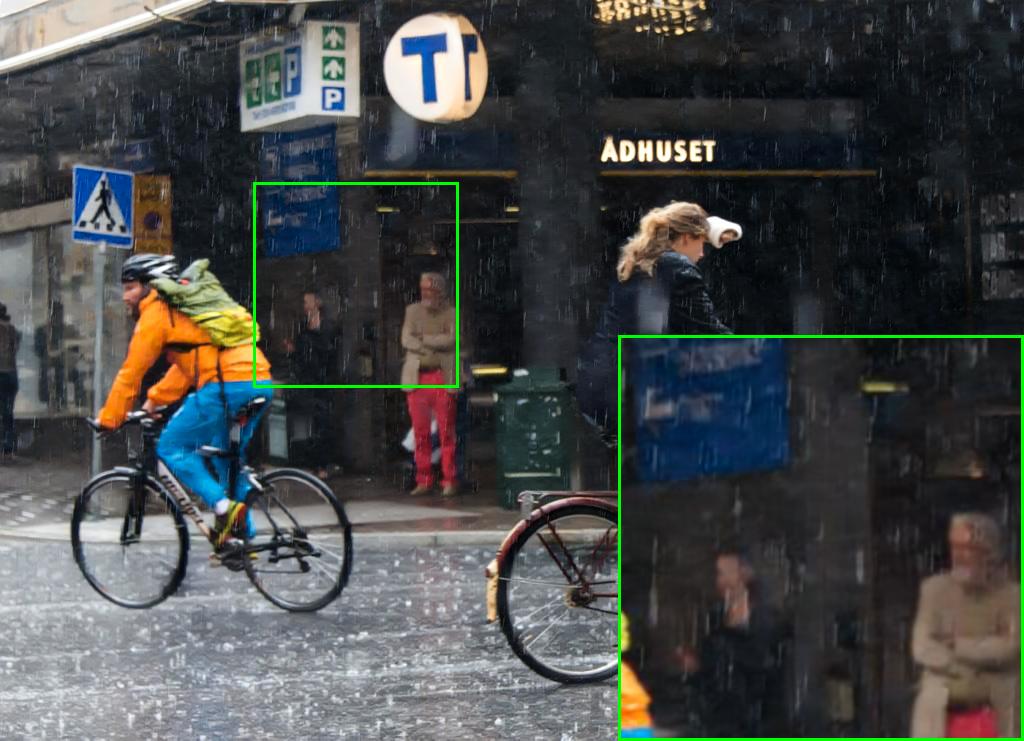}
		\scriptsize{(g) SPANet}
	\end{minipage}	
	\begin{minipage}[h]{0.117\linewidth}
		\centering
		\includegraphics[width=\linewidth]{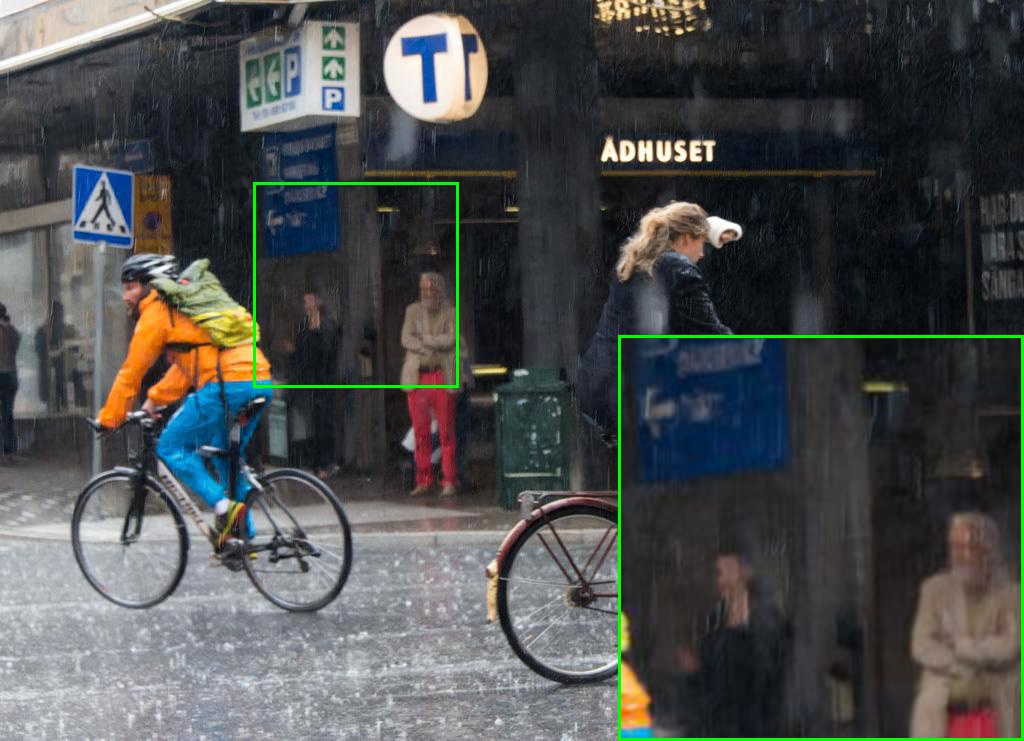}
		\scriptsize{(h) Ours}
	\end{minipage}		
	\caption{Qualitative comparisons on real rainy images.}
	\label{Fig: real_rain}
	\figvspaceloose
\end{figure*}

\begin{table}[!t]
	\caption{Quantitative comparisons of different deraining methods on benchmark~\cite{li2019single} and our synthetic dataset.}
	\centering
	\normalsize
	\begin{tabular}{|c|c|c|c|c|}
		\hline
		\multicolumn{1}{|c|}{\multirow{2}*{Method}}&\multicolumn{2}{c|}{Benchmark~\cite{li2019single}}&\multicolumn{2}{c|}{Our Dataset}\\
		\cline{2-5}
		& PSNR & SSIM & PSNR & SSIM\\    
		\hline	
		\hline
		DDN		&$23.05$	&$0.7749$ &$23.51$	&$0.7673$\\
		\hline
		JORDER 	&$27.60$	&$0.8645$ &$25.40$	&$0.8553$\\
		\hline
		DID-MDN &$29.06$	&$0.8669$ &$27.11$	&$0.8751$\\
		\hline
		PReNet 	&$30.26$	&$0.9008$ &$28.18$	&$0.8770$\\
		\hline
		RESCAN	&$29.85$	&$0.8846$ &$27.48$	&$0.8586$\\
		\hline
		SPANet	&$29.38$	&$0.9006$ &$28.20$	&$0.8983$\\
		\hline
		Ours	&$\mathbf{31.15}$	&$\mathbf{0.9131}$ &$\mathbf{30.79}$ &$\mathbf{0.9182}$\\
		\hline
	\end{tabular}
	\label{tab:rain}
	\tabvspace
\end{table}

\begin{table}[!t]
	\centering
	\caption{\label{tab:haze}Quantitative comparisons of different dehazing methods on indoor and outdoor scenes in SOTS.}
	\centering
	\normalsize
	\begin{tabular}{|c|c|c|c|c|}
		\hline
		\multirow{2}*{Methods}  &  \multicolumn{2}{c|}{Indoor}    &   \multicolumn{2}{c|}{Outdoor}\\
		\cline{2-5}
		&   PSNR    &SSIM   &PSNR   &SSIM\\
		\hline
		\hline
		DCP	&$16.61$	&$0.8546$	&$19.14$	&$0.8605$\\
		\hline
		DehazeNet &$19.82$	&$0.8209$	&$24.75$	&$0.9269$\\
		\hline
		MSCNN 	&$19.84$	&$0.8327$	&$22.06$	&$0.9078$\\
		\hline
		AOD-Net &$20.51$	&$0.8162$	&$24.14$	&$0.9198$\\
		\hline
		GFN	&$24.91$	&$0.9186$	&$28.29$	&$0.9621$\\
		\hline
		Ours &$\mathbf{34.35}$   &$\mathbf{0.9923}$   &$\mathbf{33.12}$   &$\mathbf{0.9882}$\\
		\hline
	\end{tabular}
	\tabvspace
\end{table}

\begin{figure*}[!t]
	\centering
	\begin{minipage}[h]{0.117\linewidth}
		\centering
		\includegraphics[width=\linewidth]{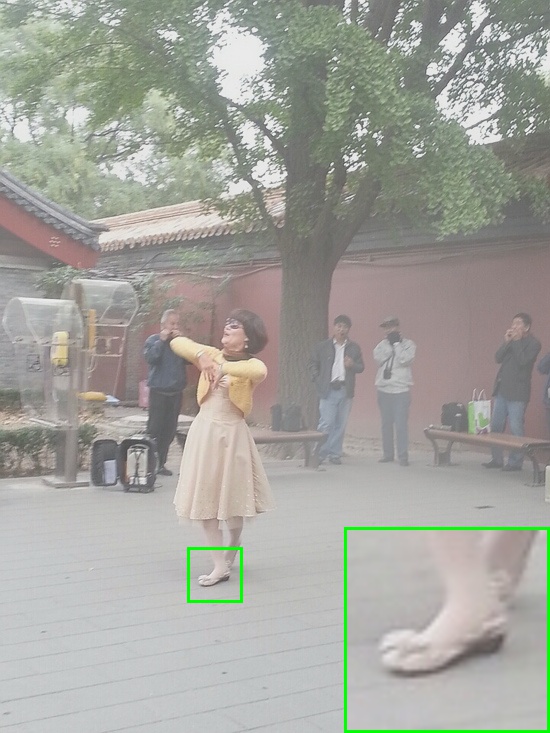}
	\end{minipage}
	\begin{minipage}[h]{0.117\linewidth}
		\centering
		\includegraphics[width=\linewidth]{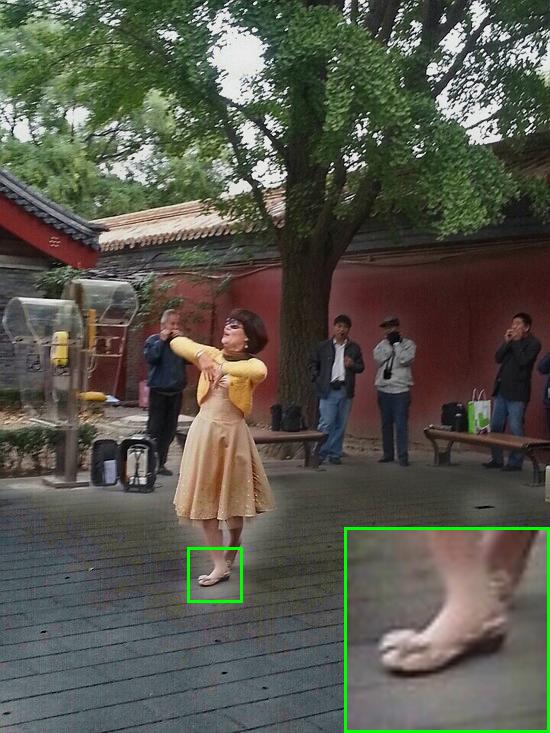}
	\end{minipage}
	\begin{minipage}[h]{0.117\linewidth}
		\centering
		\includegraphics[width=\linewidth]{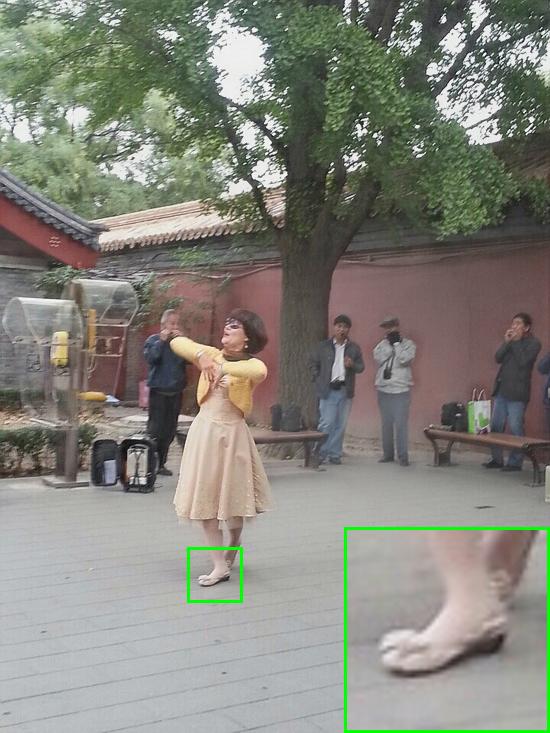}
	\end{minipage}	
	\begin{minipage}[h]{0.117\linewidth}
		\centering
		\includegraphics[width=\linewidth]{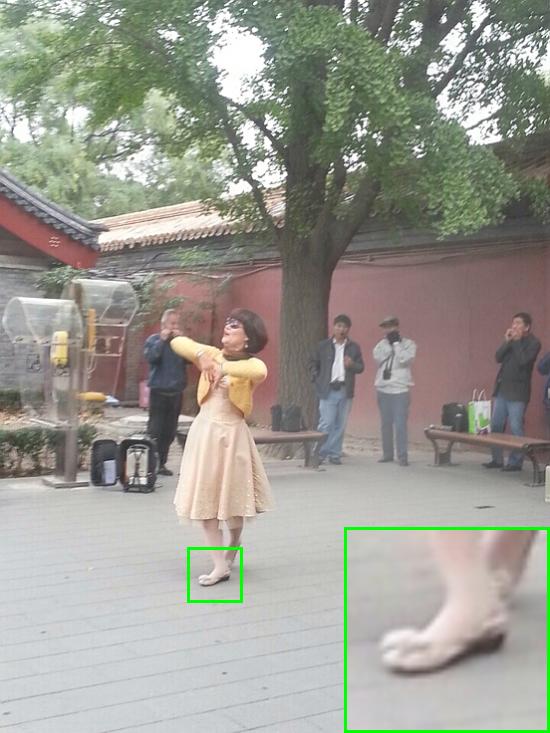}
	\end{minipage}	
	\begin{minipage}[h]{0.117\linewidth}
		\centering
		\includegraphics[width=\linewidth]{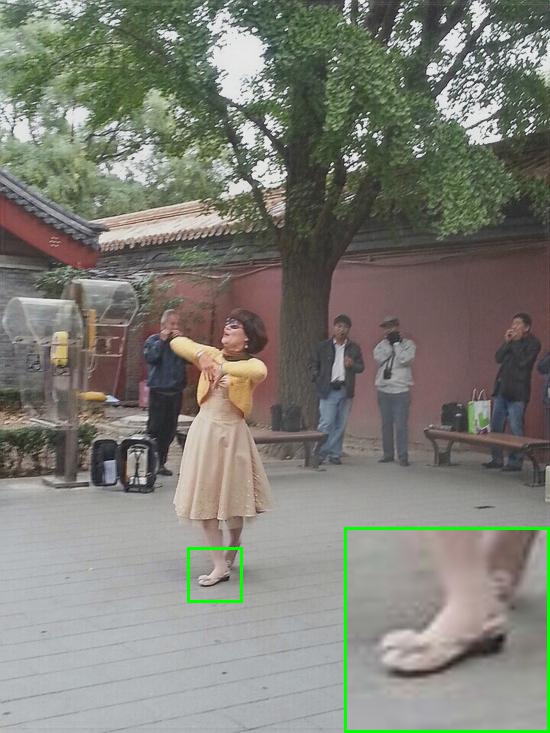}
	\end{minipage}		
	\begin{minipage}[h]{0.117\linewidth}
		\centering
		\includegraphics[width=\linewidth]{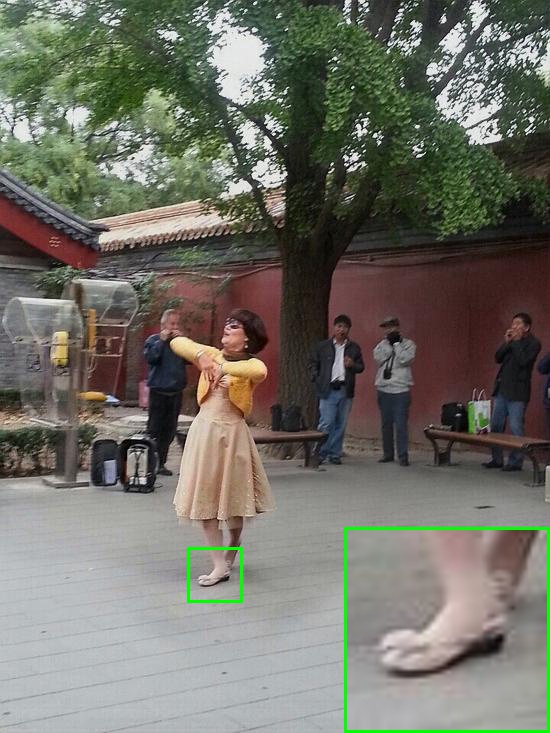}
	\end{minipage}	
	\begin{minipage}[h]{0.117\linewidth}
		\centering
		\includegraphics[width=\linewidth]{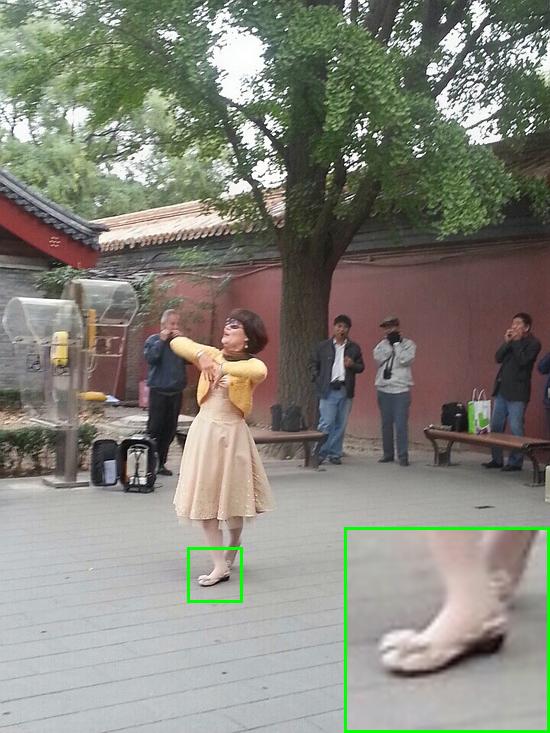}
	\end{minipage}		
	\begin{minipage}[h]{0.117\linewidth}
		\centering
		\includegraphics[width=\linewidth]{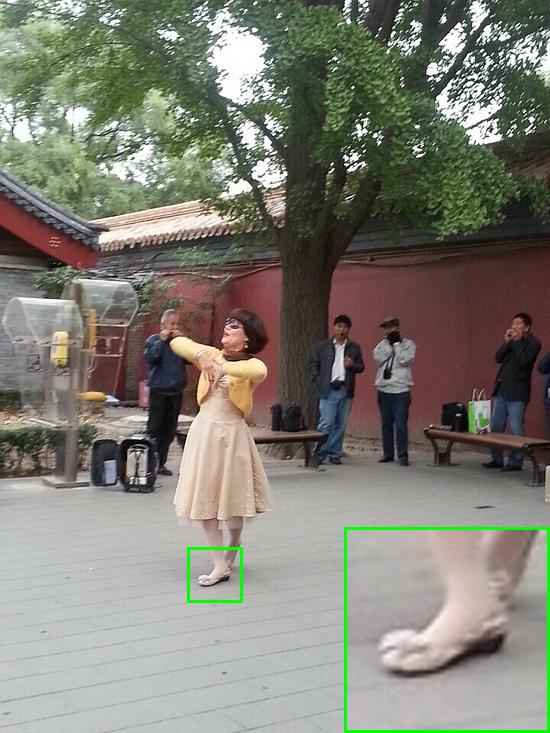}
	\end{minipage}		
	\begin{minipage}[h]{0.117\linewidth}
		\centering
		\includegraphics[width=\linewidth]{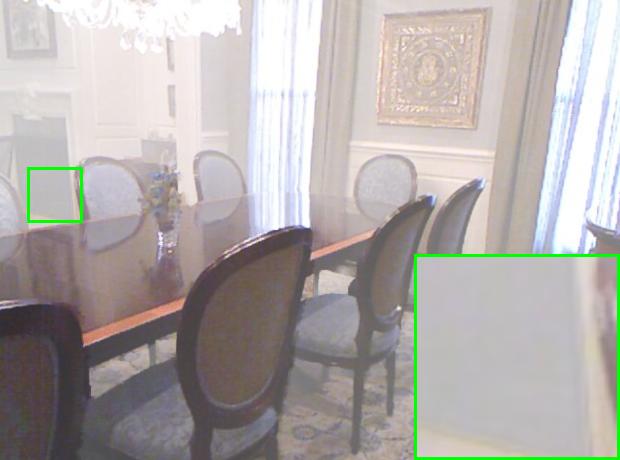}
		\scriptsize{(a) Hazy}
	\end{minipage}
	\begin{minipage}[h]{0.117\linewidth}
		\centering
		\includegraphics[width=\linewidth]{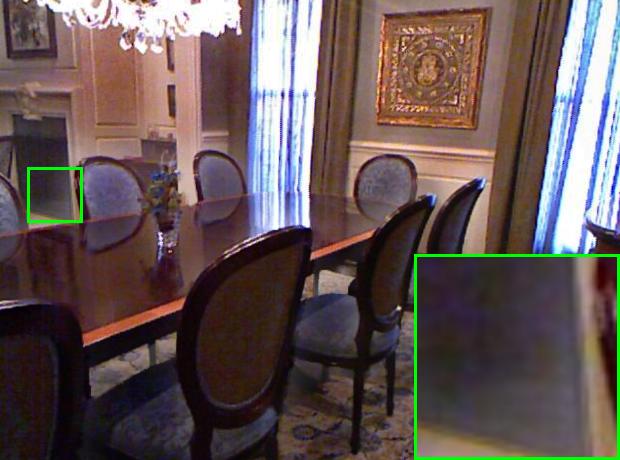}
		\scriptsize{(b) DCP}
	\end{minipage}
	\begin{minipage}[h]{0.117\linewidth}
		\centering
		\includegraphics[width=\linewidth]{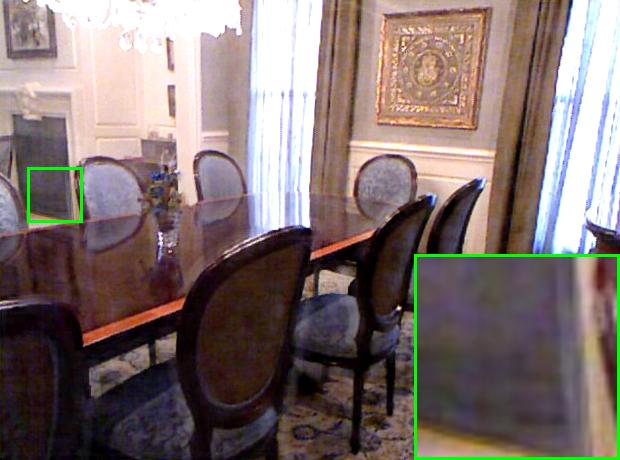}
		\scriptsize{(c) DehazeNet}
	\end{minipage}	
	\begin{minipage}[h]{0.117\linewidth}
		\centering
		\includegraphics[width=\linewidth]{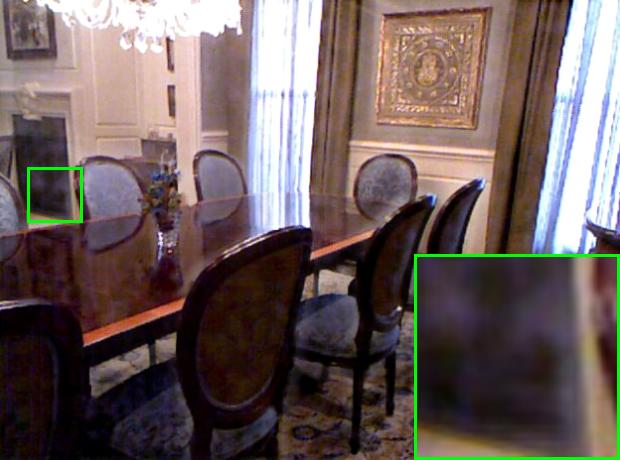}
		\scriptsize{(d) MSCNN}
	\end{minipage}	
	\begin{minipage}[h]{0.117\linewidth}
		\centering
		\includegraphics[width=\linewidth]{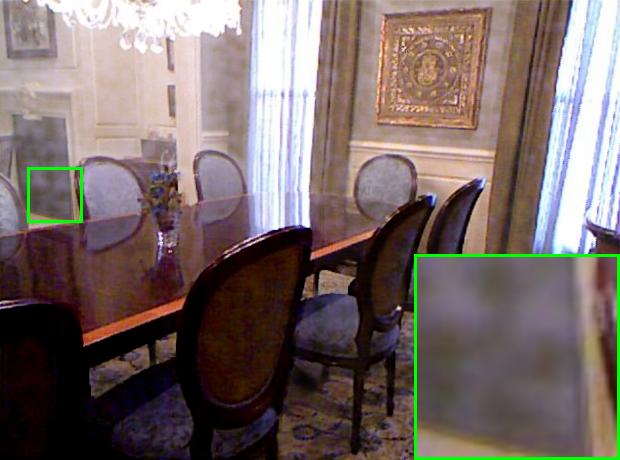}
		\scriptsize{(e) AOD-Net}
	\end{minipage}		
	\begin{minipage}[h]{0.117\linewidth}
		\centering
		\includegraphics[width=\linewidth]{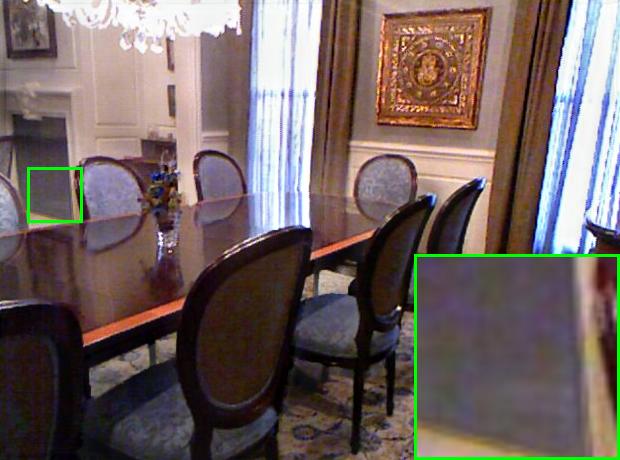}
		\scriptsize{(f) GFN}
	\end{minipage}	
	\begin{minipage}[h]{0.117\linewidth}
		\centering
		\includegraphics[width=\linewidth]{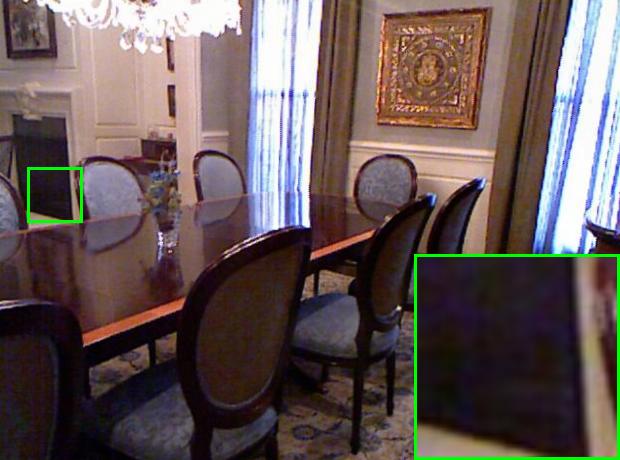}
		\scriptsize{(g) Ours}
	\end{minipage}		
	\begin{minipage}[h]{0.117\linewidth}
		\centering
		\includegraphics[width=\linewidth]{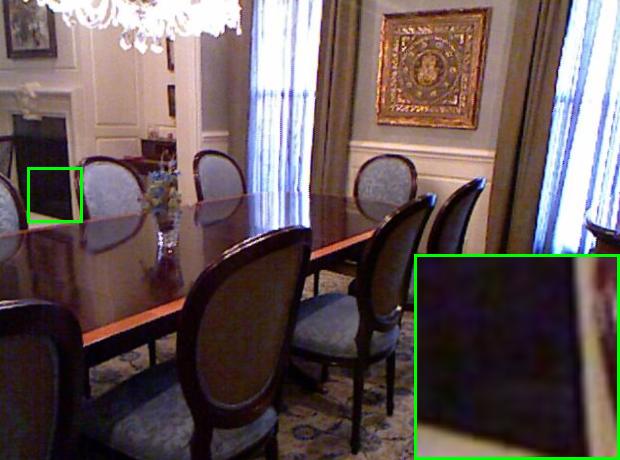}
		\scriptsize{(h) GT}
	\end{minipage}
	\caption{Qualitative comparisons on the Synthetic Objective Testing Set (SOTS)~\cite{li2019benchmarking}.}
	\label{fig:synthetic_haze}
	\figvspaceloose
\end{figure*}

\begin{figure*}[!t]
	\centering
	\begin{minipage}[h]{0.135\linewidth}
		\centering
		\includegraphics[width=\linewidth]{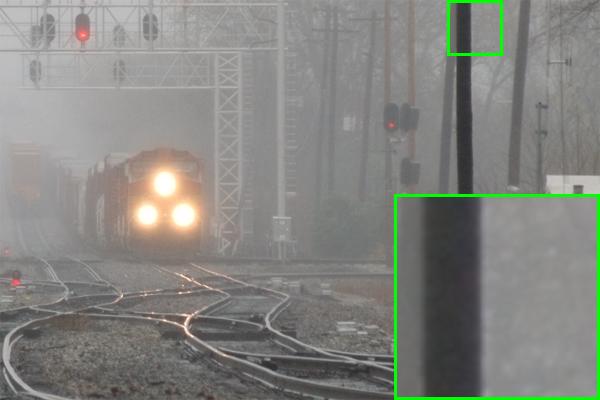}
	\end{minipage}
	\begin{minipage}[h]{0.135\linewidth}
		\centering
		\includegraphics[width=\linewidth]{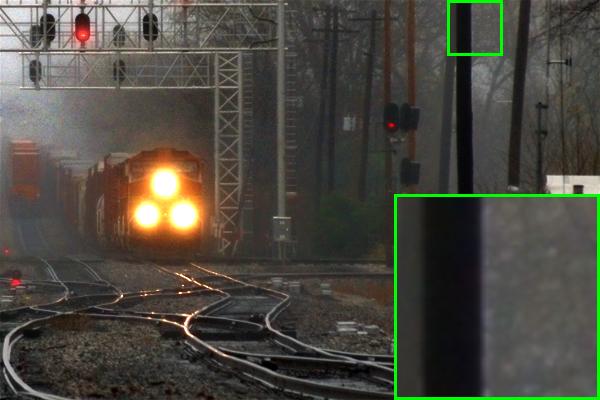}
	\end{minipage}
	\begin{minipage}[h]{0.135\linewidth}
		\centering
		\includegraphics[width=\linewidth]{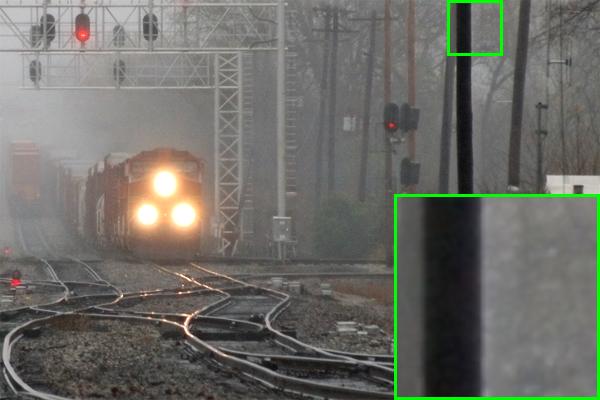}
	\end{minipage}	
	\begin{minipage}[h]{0.135\linewidth}
		\centering
		\includegraphics[width=\linewidth]{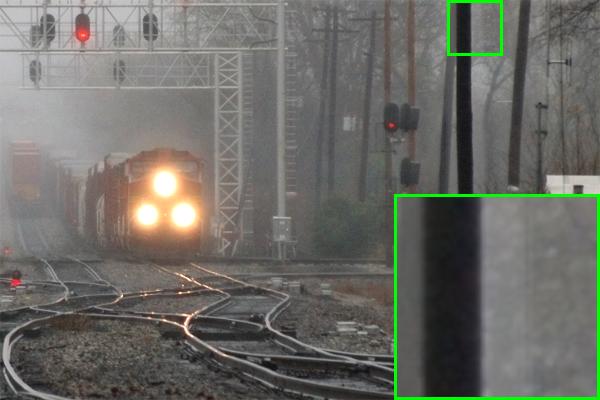}
	\end{minipage}	
	\begin{minipage}[h]{0.135\linewidth}
		\centering
		\includegraphics[width=\linewidth]{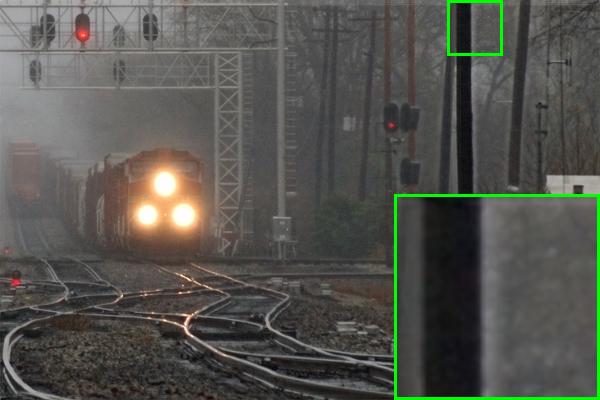}
	\end{minipage}		
	\begin{minipage}[h]{0.135\linewidth}
		\centering
		\includegraphics[width=\linewidth]{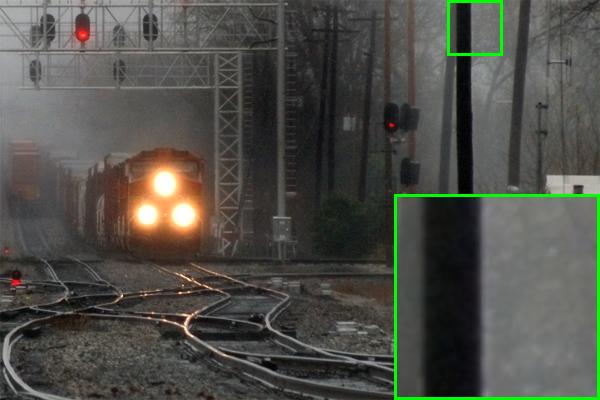}
	\end{minipage}	
	\begin{minipage}[h]{0.135\linewidth}
		\centering
		\includegraphics[width=\linewidth]{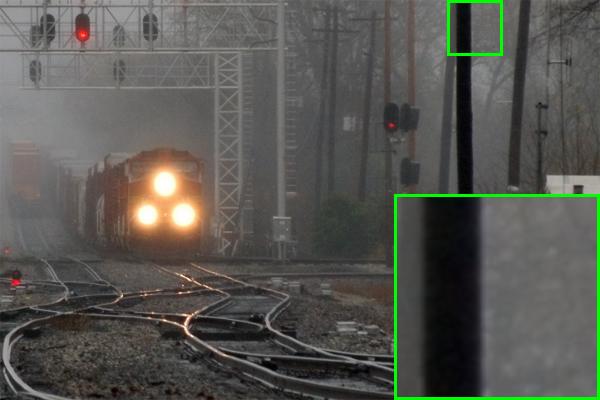}
	\end{minipage}			
	\begin{minipage}[h]{0.135\linewidth}
		\centering
		\includegraphics[width=\linewidth]{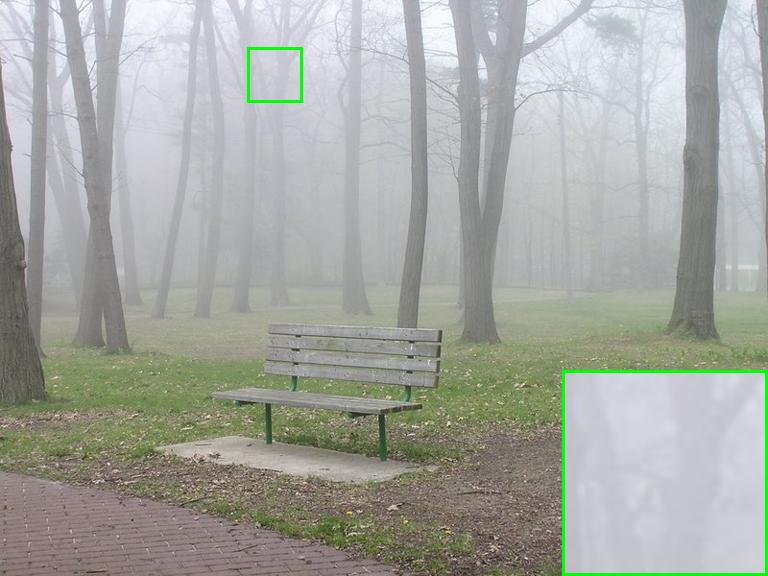}
		\scriptsize{(a) Hazy}
	\end{minipage}
	\begin{minipage}[h]{0.135\linewidth}
		\centering
		\includegraphics[width=\linewidth]{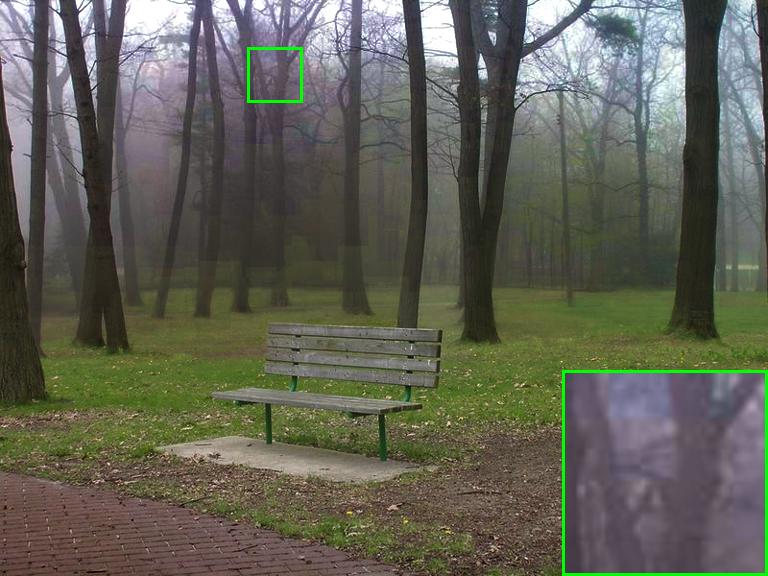}
		\scriptsize{(b) DCP}
	\end{minipage}
	\begin{minipage}[h]{0.135\linewidth}
		\centering
		\includegraphics[width=\linewidth]{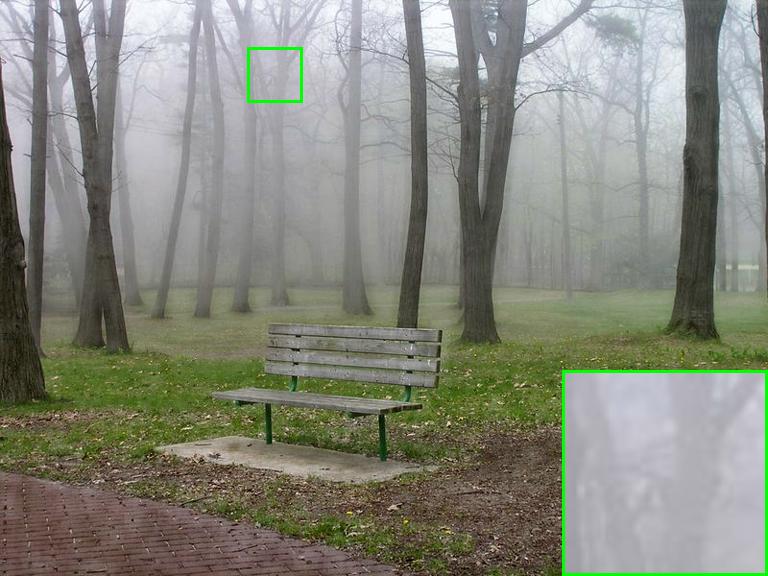}
		\scriptsize{(c) DehazeNet}
	\end{minipage}	
	\begin{minipage}[h]{0.135\linewidth}
		\centering
		\includegraphics[width=\linewidth]{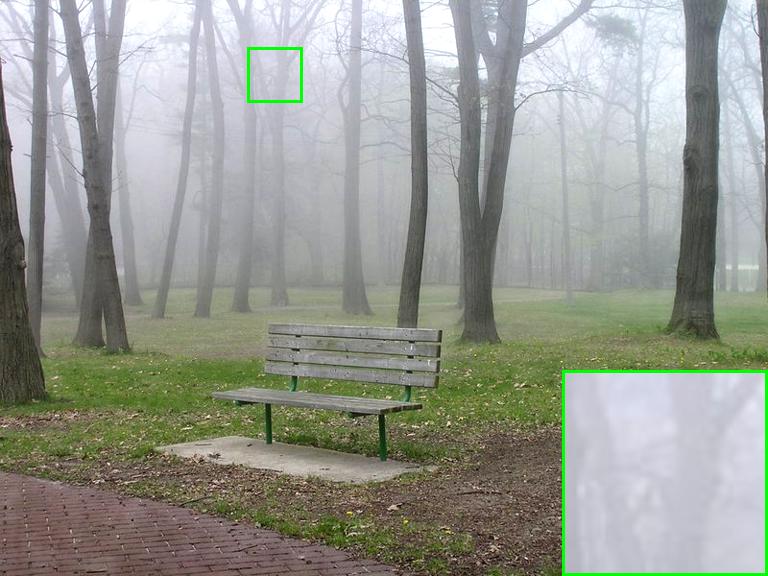}
		\scriptsize{(d) MSCNN}
	\end{minipage}	
	\begin{minipage}[h]{0.135\linewidth}
		\centering
		\includegraphics[width=\linewidth]{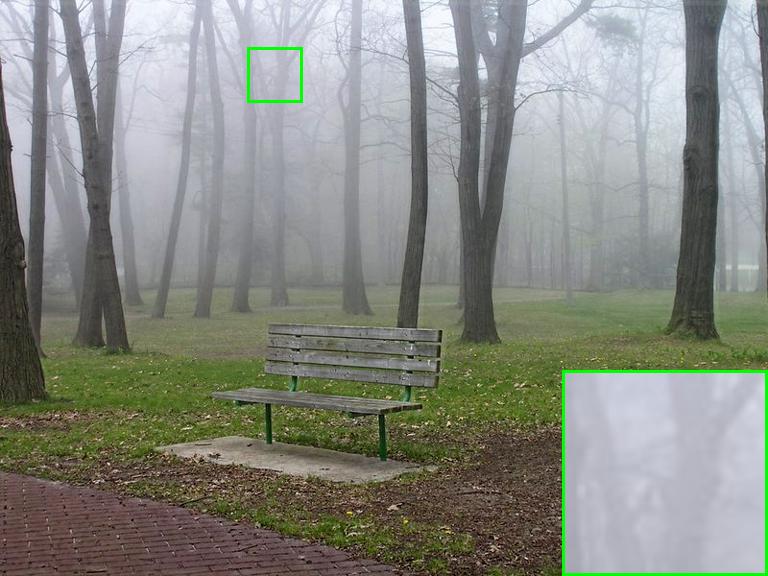}
		\scriptsize{(e) AOD-Net}
	\end{minipage}		
	\begin{minipage}[h]{0.135\linewidth}
		\centering
		\includegraphics[width=\linewidth]{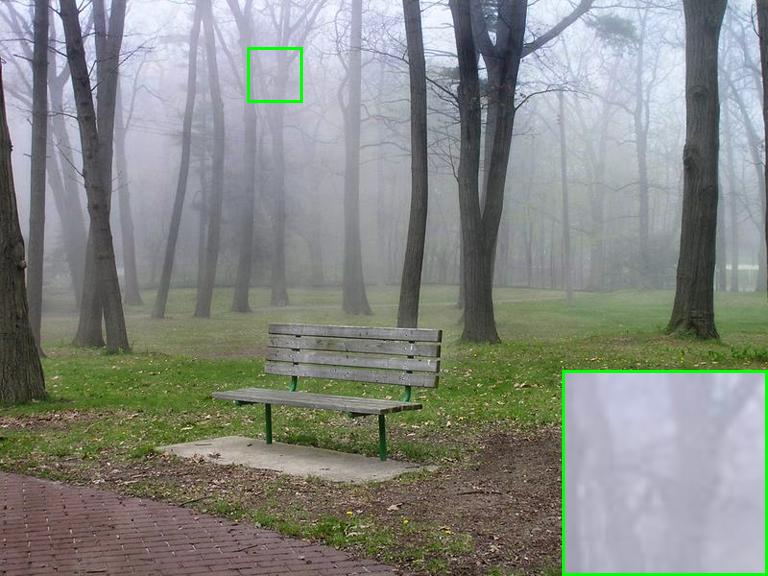}
		\scriptsize{(f) GFN}
	\end{minipage}	
	\begin{minipage}[h]{0.135\linewidth}
		\centering
		\includegraphics[width=\linewidth]{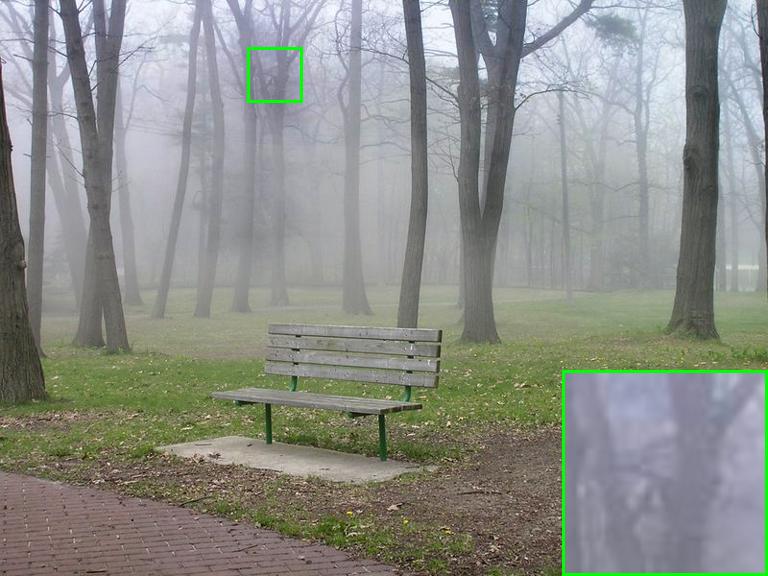}
		\scriptsize{(g) Ours}
	\end{minipage}		
	\caption{Qualitative comparisons on the real hazy images.}
	\label{fig:real_haze}
	\figvspaceloose
\end{figure*}

\subsection{Quantitative and Qualitative Comparisons}
We quantitatively and qualitatively compare the proposed DSCAN with several existing deraining and dehazing methods. Specifically, DDN~\cite{fu2017removing}, JORDER~\cite{yang2017deep}, DID-MDN~\cite{zhang2018density}, PReNet~\cite{ren2019progressive}, RESCAN~\cite{li2018recurrent}, and SPANet~\cite{wang2019spatial} are chosen as representatives of the existing deraining methods while  DCP~\cite{iebmdcp01}, DehazeNet~\cite{iebmdehazenet01}, MSCNN~\cite{iebmmscnn01}, AOD-Net~\cite{iebmaod01}, and GFN~\cite{iebmgated01} are chosen as representatives of the existing dehazing methods. Except for DCP, all the aforementioned methods are deep-learning-based. Moreover,  SPANet and GFN can be considered respectively as the current state-of-the-art for deraining and dehazing. For fair comparisons, we laboriously retrain all the methods under consideration using the same training strategy described in Section \ref{implementation}. 

The quantitative comparisons of different deraining methods are provided in Table~\ref{tab:rain} while the qualitative comparisons are illustrated in Fig.~\ref{fig:synthetic_rain} (synthetic images) and Fig.~\ref{Fig: real_rain} (real images). As to the dehazing methods, we show the quantitative comparisons in Table~\ref{tab:haze} and the  qualitative comparisons in Fig.~\ref{fig:synthetic_haze} (synthetic images) and Fig.~\ref{fig:real_haze} (real images). For both tasks, it is apparent that the proposed DSCAN outperforms all other methods under comparison. For synthetic images, our method delivers the visually clearest deraining/dehazing  results (see, e.g., the pole in Fig.~\ref{fig:synthetic_rain} and the closet area in Fig.~\ref{fig:synthetic_haze}); our method also succeeds in suppressing veiling and halo artifacts usually caused by incomplete removal of rain/haze effects (see, e.g.,  the street in Fig.~\ref{fig:synthetic_rain} and the foot in Fig.~\ref{fig:synthetic_haze}). Moreover, the proposed DSCAN produces appealing results on real rainy images by removing most of rain streaks, whereas the derained results by the other methods still suffer from visual degradation due to the residual rain effect; for real hazy images, the results by our method are  free of major color distortions and halo artifacts, and thus performs favorably against the others. It can also be seen from Table~\ref{tab:rain} that the superior performance of the proposed DSCAN is not confined to a specific dataset.

\begin{table}[t]
	\centering
	\caption{Dual-task DSCAN with joint learning on RESIDE and our synthetic dataset of rainy images.}
	\centering
	\normalsize
	\begin{tabular}{|c|c|c|c|c|}
		\hline
		{Test data}& PSNR & SSIM\\    
		\hline	
		\hline
		RESIDE outdoor~\cite{li2019benchmarking}	&$30.10$	&$0.9685$\\
		\hline
		Our rainy dataset &$30.35$  &$0.9160$\\
		\hline	
	\end{tabular}
	\tabvspace
	\label{tab:jointlearning}
\end{table}

\begin{figure}[!b]
	\centering
	\begin{minipage}[h]{0.49\linewidth}
		\centering
		\includegraphics[width=\linewidth]{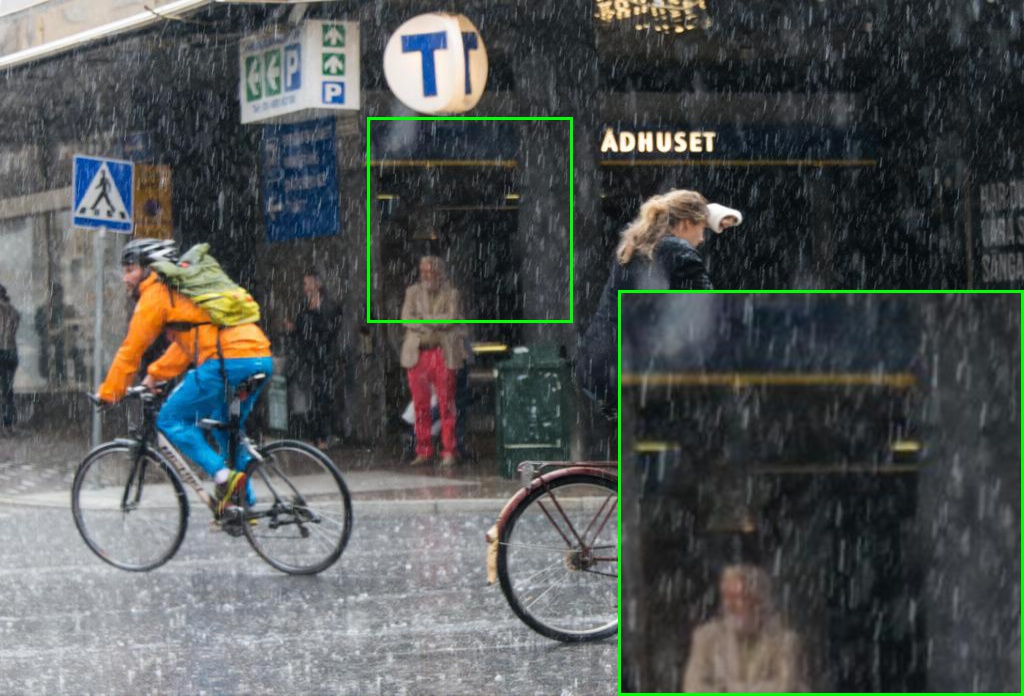}
	\end{minipage}
	\begin{minipage}[h]{0.49\linewidth}
		\centering
		\includegraphics[width=\linewidth]{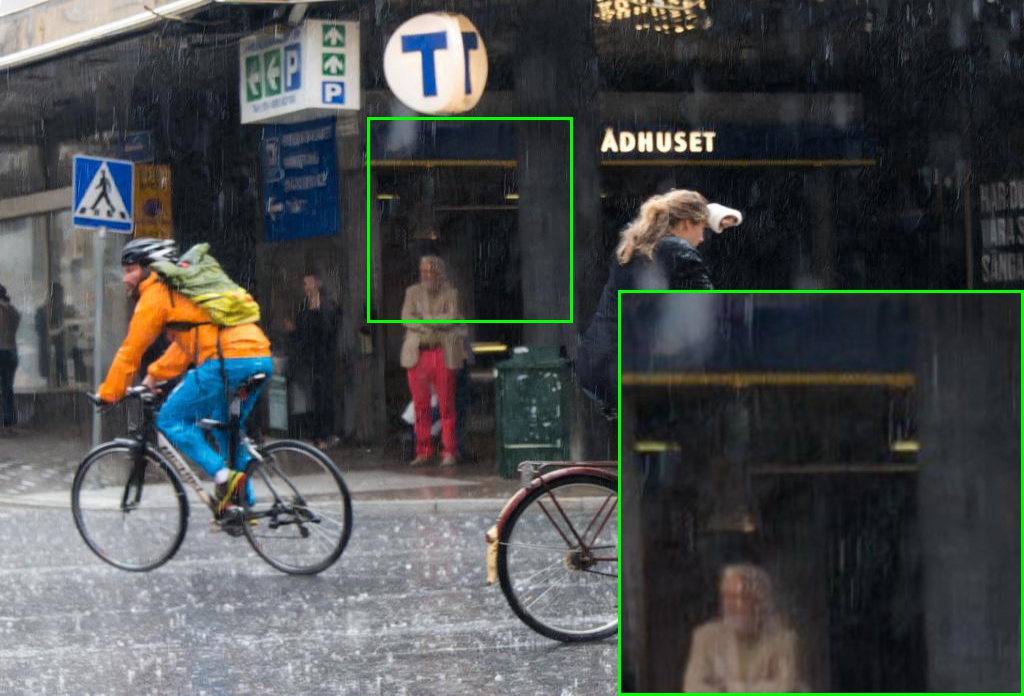}
	\end{minipage}
	\\
	\begin{minipage}[h]{0.49\linewidth}
		\centering
		\includegraphics[width=\linewidth]{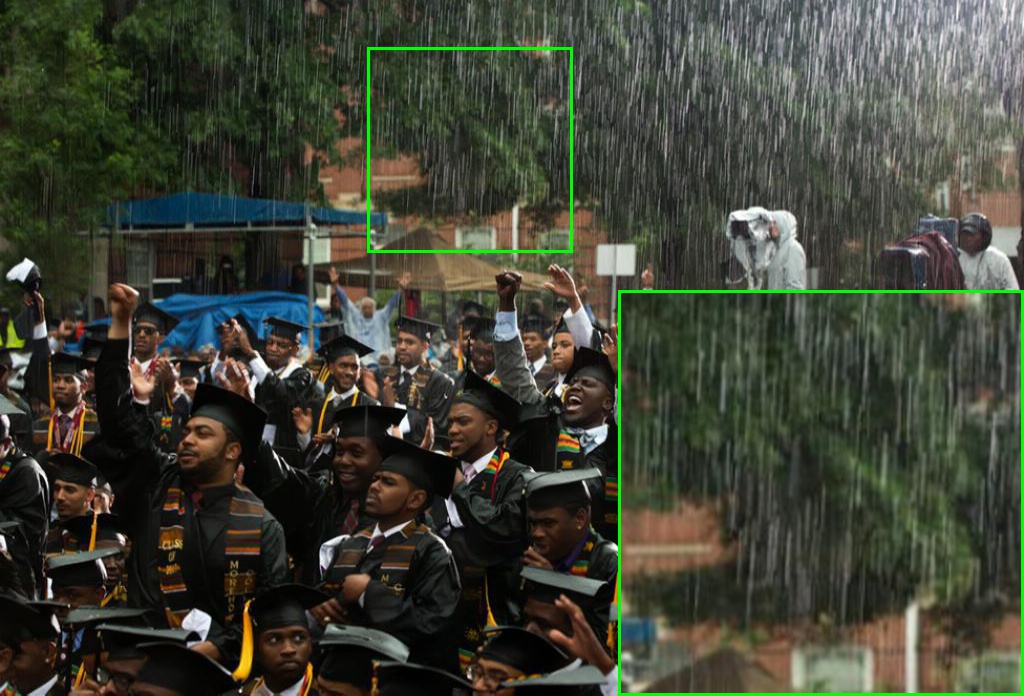}
		\scriptsize{(a) Benchmark~\cite{li2019single}}
	\end{minipage}		
	\begin{minipage}[h]{0.49\linewidth}
		\centering
		\includegraphics[width=\linewidth]{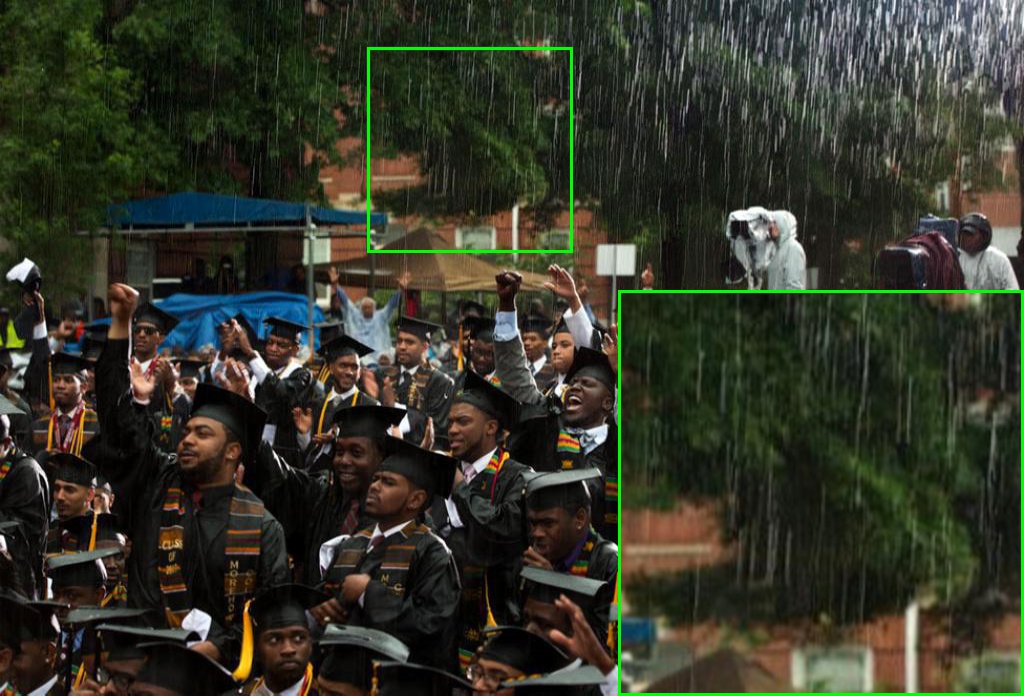}
		\scriptsize{(b) Ours}
	\end{minipage}	
	\caption{Qualitative comparisons of the generalization performance of two versions of DSCAN on real rainy images, one trained using Benchmark~\cite{li2019single} and the other using the new synthetic dataset.}
	\label{fig:differentweights}
	\figvspaceloose
\end{figure}

So far the deraining methods and the dehazing methods are trained on different datasets. However, the proposed DSCAN can actually capitalize on both types of datasets to learn deraining and dehazing simultaneously.  To demonstrate this, we combine RESIDE and our synthetic dataset of rainy images, and use the combined version to train a dual-task DSCAN. It can be seen from Table~\ref{tab:jointlearning} that the dual-task DSCAN suffers from some performance degradation as compared to its single-task counterparts; nevertheless, it still outperforms the other deraining and dehazing methods under comparison.

To gain more insight into different rain models, 
we train two versions of the proposed DSCAN, one using benchmark~\cite{li2019single} and the other using the new synthetic dataset based on our physical model, and test them on real rainy images. It can be seen from Fig.~\ref{fig:differentweights} that the version trained using the new synthetic dataset has superior generalization performance. This shows that our physical model better captures the real rain effect as compared to the over-simplified model (\ref{equ:basic}) adopted by benchmark~\cite{li2019single}.

\subsection{Ablation Study}

We conduct various ablation studies to justify the overall design of the proposed DSCAN. Table~\ref{tab:ablation} shows  the quantitative comparisons of DSCAN and its three variants: 1) removing the dense connections across different scales (w/o dense-connection), 2) performing direct feature addition without channel-wise attention (w/o attentive addition), 3) replacing the proposed CSA-RDB with the original version of RDB (w/ RDB). The comparison results provide strong evidences in favor of the proposed design.

\begin{table}[!t]
	\caption{\label{tab:ablation}Ablation studies for the proposed DSCAN.}
	\centering
	\normalsize
	\begin{tabular}{|c|c|c|c|c|}
		\hline
		{Method}& PSNR & SSIM\\    
		\hline	
		\hline
		w/o dense-connection	&$28.26$	&$0.8948$\\
		\hline
		w/o attentive addition &$30.46$  &$0.9138$\\
		\hline
		w/ RDB &$30.06$	&$0.9095$\\
		\hline
		Ours	&$\mathbf{30.79}$	&$\mathbf{0.9182}$\\
		\hline	
	\end{tabular}
	\tabvspace
\end{table}

\subsection{Runtime Analysis}
\begin{figure}[h]
	\centering
	\begin{minipage}[h]{0.8\linewidth}
		\centering
		\includegraphics[width=\linewidth]{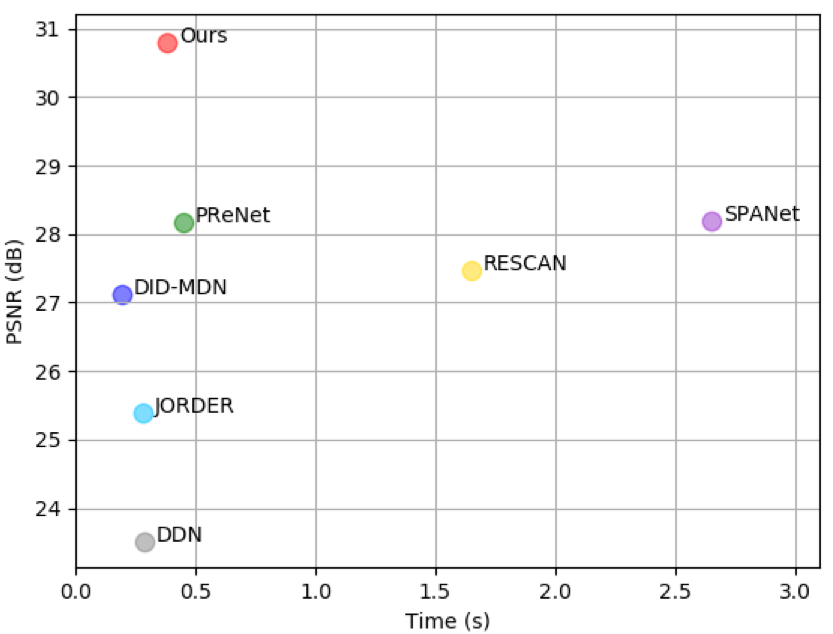}
		\scriptsize{(a) Deraining}
	\end{minipage}
	\hspace{0.5cm}
	\begin{minipage}[h]{0.8\linewidth}
		\centering
		\includegraphics[width=\linewidth]{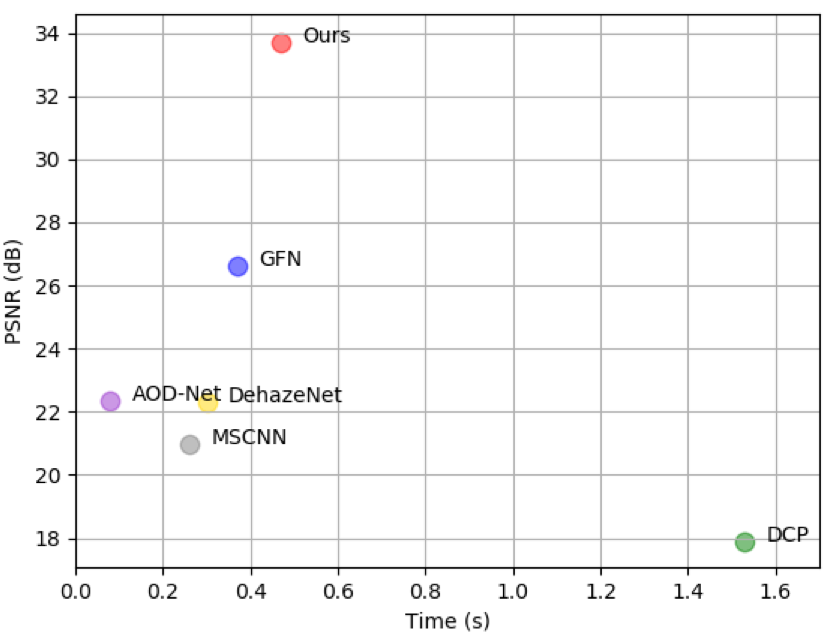}
		\scriptsize{(b) Dehazing}
	\end{minipage}
	\caption{Runtime vs.\ PSNR for different deraining/dehazing methods.}
	\label{fig:runtime}
\end{figure}

Fig.~\ref{fig:runtime} plots runtime vs.\ PSNR for different deraining/dehazing methods. Here runtime refers to the average time of processing one rainy/hazy test image. It can be seen that the proposed DSCAN achieves the highest PSNR value with competitive runtime in both tasks.

\section{Conclusion}

In this paper, we have developed a unified physical model for rain and haze effects. This model is, to a large extent, generic in nature, and can potentially accommodate other weather conditions as well. The proposed DSCAN is inspired by our study of multi-scale information exchange and aggregation, and is likely to find applications beyond deraining and dehazing. 
In this sense, our work can be viewed as a step forward in theory building and algorithm design for general-purpose deweathering.

\bibliographystyle{IEEEtran}
\bibliography{IEEEabrv,egbib}

\begin{thebibliography}{10}
\providecommand{\url}[1]{#1}
\csname url@samestyle\endcsname
\providecommand{\newblock}{\relax}
\providecommand{\bibinfo}[2]{#2}
\providecommand{\BIBentrySTDinterwordspacing}{\spaceskip=0pt\relax}
\providecommand{\BIBentryALTinterwordstretchfactor}{4}
\providecommand{\BIBentryALTinterwordspacing}{\spaceskip=\fontdimen2\font plus
\BIBentryALTinterwordstretchfactor\fontdimen3\font minus
  \fontdimen4\font\relax}
\providecommand{\BIBforeignlanguage}[2]{{%
\expandafter\ifx\csname l@#1\endcsname\relax
\typeout{** WARNING: IEEEtran.bst: No hyphenation pattern has been}%
\typeout{** loaded for the language `#1'. Using the pattern for}%
\typeout{** the default language instead.}%
\else
\language=\csname l@#1\endcsname
\fi
#2}}
\providecommand{\BIBdecl}{\relax}
\BIBdecl

\bibitem{bijelic2018benchmarking}
M.~Bijelic, T.~Gruber, and W.~Ritter, ``Benchmarking image sensors under
  adverse weather conditions for autonomous driving,'' in \emph{Proceedings of
  the IEEE Intelligent Vehicles Symposium}, 2018, pp. 1773--1779.

\bibitem{kong2017millimeter}
L.~Kong, M.~K. Khan, F.~Wu, G.~Chen, and P.~Zeng, ``Millimeter-wave wireless
  communications for iot-cloud supported autonomous vehicles: Overview, design,
  and challenges,'' \emph{IEEE Communications Magazine}, vol.~55, no.~1, pp.
  62--68, 2017.

\bibitem{fu2017removing}
X.~Fu, J.~Huang, D.~Zeng, Y.~Huang, X.~Ding, and J.~Paisley, ``Removing rain
  from single images via a deep detail network,'' in \emph{Proceedings of the
  IEEE Conference on Computer Vision and Pattern Recognition}, 2017, pp.
  3855--3863.

\bibitem{zhang2018density}
H.~Zhang and V.~M. Patel, ``Density-aware single image de-raining using a
  multi-stream dense network,'' in \emph{Proceedings of the IEEE Conference on
  Computer Vision and Pattern Recognition}, 2018, pp. 695--704.

\bibitem{ren2019progressive}
D.~Ren, W.~Zuo, Q.~Hu, P.~Zhu, and D.~Meng, ``Progressive image deraining
  networks: a better and simpler baseline,'' in \emph{Proceedings of the IEEE
  Conference on Computer Vision and Pattern Recognition}, 2019, pp. 3937--3946.

\bibitem{iebmmscnn01}
W.~Ren, S.~Liu, H.~Zhang, J.~Pan, X.~Cao, and M.-H. Yang, ``Single image
  dehazing via multi-scale convolutional neural networks,'' in
  \emph{Proceedings of the European conference on computer vision}.\hskip 1em
  plus 0.5em minus 0.4em\relax Springer, 2016, pp. 154--169.

\bibitem{wang2019spatial}
T.~Wang, X.~Yang, K.~Xu, S.~Chen, Q.~Zhang, and R.~W. Lau, ``Spatial attentive
  single-image deraining with a high quality real rain dataset,'' in
  \emph{Proceedings of the IEEE Conference on Computer Vision and Pattern
  Recognition}, 2019, pp. 12\,270--12\,279.

\bibitem{iebmaod01}
B.~Li, X.~Peng, Z.~Wang, J.~Xu, and D.~Feng, ``Aod-net: All-in-one dehazing
  network,'' in \emph{Proceedings of the IEEE International Conference on
  Computer Vision}, 2017, pp. 4770--4778.

\bibitem{iebmgated01}
W.~Ren, L.~Ma, J.~Zhang, J.~Pan, X.~Cao, W.~Liu, and M.-H. Yang, ``Gated fusion
  network for single image dehazing,'' in \emph{Proceedings of the IEEE
  Conference on Computer Vision and Pattern Recognition}, 2018, pp. 3253--3261.

\bibitem{mccartney1976optics}
E.~J. McCartney, ``Optics of the atmosphere: scattering by molecules and
  particles,'' \emph{New York, John Wiley and Sons, Inc., 1976. 421 p.}, 1976.

\bibitem{narasimhan2000chromatic}
S.~G. Narasimhan and S.~K. Nayar, ``Chromatic framework for vision in bad
  weather,'' in \emph{Proceedings IEEE Conference on Computer Vision and
  Pattern Recognition}, vol.~1.\hskip 1em plus 0.5em minus 0.4em\relax IEEE,
  2000, pp. 598--605.

\bibitem{narasimhan2002vision}
------, ``Vision and the atmosphere,'' \emph{International Journal of Computer
  Vision}, vol.~48, no.~3, pp. 233--254, 2002.

\bibitem{yang2017deep}
W.~Yang, R.~T. Tan, J.~Feng, J.~Liu, Z.~Guo, and S.~Yan, ``Deep joint rain
  detection and removal from a single image,'' in \emph{Proceedings of the IEEE
  Conference on Computer Vision and Pattern Recognition}, 2017, pp. 1357--1366.

\bibitem{li2019single}
S.~Li, I.~B. Araujo, W.~Ren, Z.~Wang, E.~K. Tokuda, R.~H. Junior,
  R.~Cesar-Junior, J.~Zhang, X.~Guo, and X.~Cao, ``Single image deraining: A
  comprehensive benchmark analysis,'' in \emph{Proceedings of the IEEE
  Conference on Computer Vision and Pattern Recognition}, 2019, pp. 3838--3847.

\bibitem{zhu2017unpaired}
J.-Y. Zhu, T.~Park, P.~Isola, and A.~A. Efros, ``Unpaired image-to-image
  translation using cycle-consistent adversarial networks,'' in
  \emph{Proceedings of the IEEE International Conference on Computer Vision},
  2017, pp. 2223--2232.

\bibitem{goodfellow2014generative}
I.~Goodfellow, J.~Pouget-Abadie, M.~Mirza, B.~Xu, D.~Warde-Farley, S.~Ozair,
  A.~Courville, and Y.~Bengio, ``Generative adversarial nets,'' in
  \emph{Advances in Neural Information Processing Systems}, 2014, pp.
  2672--2680.

\bibitem{shen2018deep}
Z.~Shen, W.-S. Lai, T.~Xu, J.~Kautz, and M.-H. Yang, ``Deep semantic face
  deblurring,'' in \emph{Proceedings of the IEEE Conference on Computer Vision
  and Pattern Recognition}, 2018, pp. 8260--8269.

\bibitem{chen2018learning}
C.~Chen, Q.~Chen, J.~Xu, and V.~Koltun, ``Learning to see in the dark,'' in
  \emph{Proceedings of the IEEE Conference on Computer Vision and Pattern
  Recognition}, 2018, pp. 3291--3300.

\bibitem{multiimagepolar01}
Y.~Y. Schechner, S.~G. Narasimhan, and S.~K. Nayar, ``Instant dehazing of
  images using polarization,'' in \emph{Proceedings of the IEEE Conference on
  Computer Vision and Pattern Recognition}, 2001, pp. 325--332.

\bibitem{multiimagepolar02}
S.~Shwartz, E.~Namer, and Y.~Y. Schechner, ``Blind haze separation,'' in
  \emph{Proceedings of the IEEE Conference on Computer Vision and Pattern
  Recognition}, vol.~2, 2006, pp. 1984--1991.

\bibitem{scatteringfn02}
S.~G. Narasimhan and S.~K. Nayar, ``Chromatic framework for vision in bad
  weather,'' in \emph{Proceedings of the IEEE Conference on Computer Vision and
  Pattern Recognition}, vol.~1, 2000, pp. 598--605.

\bibitem{multiimageweather02}
------, ``Contrast restoration of weather degraded images,'' \emph{IEEE
  Transactions on Pattern Analysis and Machine Intelligence}, no.~6, pp.
  713--724, 2003.

\bibitem{iebmcontrast01}
R.~T. Tan, ``Visibility in bad weather from a single image,'' in
  \emph{Proceedings of the IEEE Conference on Computer Vision and Pattern
  Recognition}, 2008, pp. 1--8.

\bibitem{iebmalbedo01}
R.~Fattal, ``Single image dehazing,'' \emph{ACM Transactions on Graphics},
  vol.~27, no.~3, p.~72, 2008.

\bibitem{iebmdcp01}
K.~He, J.~Sun, and X.~Tang, ``Single image haze removal using dark channel
  prior,'' \emph{IEEE Transactions on Pattern Analysis and Machine
  Intelligence}, vol.~33, no.~12, pp. 2341--2353, 2011.

\bibitem{iebmrf01}
K.~Tang, J.~Yang, and J.~Wang, ``Investigating haze-relevant features in a
  learning framework for image dehazing,'' in \emph{Proceedings of the IEEE
  Conference on Computer Vision and Pattern Recognition}, 2014, pp. 2995--3000.

\bibitem{iebmlinear01}
Q.~Zhu, J.~Mai, and L.~Shao, ``A fast single image haze removal algorithm using
  color attenuation prior,'' \emph{IEEE Transactions on Image Processing},
  vol.~24, no.~11, pp. 3522--3533, 2015.

\bibitem{iebmdehazenet01}
B.~Cai, X.~Xu, K.~Jia, C.~Qing, and D.~Tao, ``Dehazenet: An end-to-end system
  for single image haze removal,'' \emph{IEEE Transactions on Image
  Processing}, vol.~25, no.~11, pp. 5187--5198, 2016.

\bibitem{ren2017video}
W.~Ren, J.~Tian, Z.~Han, A.~Chan, and Y.~Tang, ``Video desnowing and deraining
  based on matrix decomposition,'' in \emph{Proceedings of the IEEE Conference
  on Computer Vision and Pattern Recognition}, 2017, pp. 4210--4219.

\bibitem{santhaseelan2015utilizing}
V.~Santhaseelan and V.~K. Asari, ``Utilizing local phase information to remove
  rain from video,'' \emph{International Journal of Computer Vision}, vol. 112,
  no.~1, pp. 71--89, 2015.

\bibitem{jiang2017novel}
T.-X. Jiang, T.-Z. Huang, X.-L. Zhao, L.-J. Deng, and Y.~Wang, ``A novel
  tensor-based video rain streaks removal approach via utilizing
  discriminatively intrinsic priors,'' in \emph{Proceedings of the IEEE
  Conference on Computer Vision and Pattern Recognition}, 2017, pp. 4057--4066.

\bibitem{barnum2010analysis}
P.~C. Barnum, S.~Narasimhan, and T.~Kanade, ``Analysis of rain and snow in
  frequency space,'' \emph{International Journal of Computer Vision}, vol.~86,
  no. 2-3, p. 256, 2010.

\bibitem{zheng2013single}
X.~Zheng, Y.~Liao, W.~Guo, X.~Fu, and X.~Ding, ``Single-image-based rain and
  snow removal using multi-guided filter,'' in \emph{International Conference
  on Neural Information Processing}.\hskip 1em plus 0.5em minus 0.4em\relax
  Springer, 2013, pp. 258--265.

\bibitem{li2016rain}
Y.~Li, R.~T. Tan, X.~Guo, J.~Lu, and M.~S. Brown, ``Rain streak removal using
  layer priors,'' in \emph{Proceedings of the IEEE Conference on Computer
  Vision and Pattern Recognition}, 2016, pp. 2736--2744.

\bibitem{li2018recurrent}
X.~Li, J.~Wu, Z.~Lin, H.~Liu, and H.~Zha, ``Recurrent squeeze-and-excitation
  context aggregation net for single image deraining,'' in \emph{Proceedings of
  the European Conference on Computer Vision}, 2018, pp. 254--269.

\bibitem{qian2018attentive}
R.~Qian, R.~T. Tan, W.~Yang, J.~Su, and J.~Liu, ``Attentive generative
  adversarial network for raindrop removal from a single image,'' in
  \emph{Proceedings of the IEEE Conference on Computer Vision and Pattern
  Recognition}, 2018, pp. 2482--2491.

\bibitem{meng2018removal}
L.~Meng~Tang, L.~Hong~Lim, and P.~Siebert, ``Removal of visual disruption
  caused by rain using cycle-consistent generative adversarial networks,'' in
  \emph{Proceedings of the European Conference on Computer Vision}, 2018.

\bibitem{li2019heavy}
R.~Li, L.-F. Cheong, and R.~T. Tan, ``Heavy rain image restoration: Integrating
  physics model and conditional adversarial learning,'' in \emph{Proceedings of
  the IEEE Conference on Computer Vision and Pattern Recognition}, 2019, pp.
  1633--1642.

\bibitem{zhang2019image}
H.~Zhang, V.~Sindagi, and V.~M. Patel, ``Image de-raining using a conditional
  generative adversarial network,'' \emph{IEEE Transactions on Circuits and
  Systems for Video Technology}, 2019.

\bibitem{eigen2013restoring}
D.~Eigen, D.~Krishnan, and R.~Fergus, ``Restoring an image taken through a
  window covered with dirt or rain,'' in \emph{Proceedings of the IEEE
  International Conference on Computer Vision}, 2013, pp. 633--640.

\bibitem{fu2017clearing}
X.~Fu, J.~Huang, X.~Ding, Y.~Liao, and J.~Paisley, ``Clearing the skies: A deep
  network architecture for single-image rain removal,'' \emph{IEEE Transactions
  on Image Processing}, vol.~26, no.~6, pp. 2944--2956, 2017.

\bibitem{cai2019toward}
J.~Cai, H.~Zeng, H.~Yong, Z.~Cao, and L.~Zhang, ``Toward real-world single
  image super-resolution: A new benchmark and a new model,'' \emph{arXiv
  preprint arXiv:1904.00523}, 2019.

\bibitem{ronneberger2015u}
O.~Ronneberger, P.~Fischer, and T.~Brox, ``U-net: Convolutional networks for
  biomedical image segmentation,'' in \emph{International Conference on Medical
  Image Computing and Computer-Assisted Intervention}.\hskip 1em plus 0.5em
  minus 0.4em\relax Springer, 2015, pp. 234--241.

\bibitem{iebmrdn01}
Y.~Zhang, Y.~Tian, Y.~Kong, B.~Zhong, and Y.~Fu, ``Residual dense network for
  image super-resolution,'' in \emph{Proceedings of the IEEE Conference on
  Computer Vision and Pattern Recognition}, 2018, pp. 2472--2481.

\bibitem{hu2018squeeze}
J.~Hu, L.~Shen, and G.~Sun, ``Squeeze-and-excitation networks,'' in
  \emph{Proceedings of the IEEE Conference on Computer Vision and Pattern
  Recognition}, 2018, pp. 7132--7141.

\bibitem{nair2010rectified}
V.~Nair and G.~E. Hinton, ``Rectified linear units improve restricted boltzmann
  machines,'' in \emph{Proceedings of the 27th International Conference on
  Machine Learning}, 2010, pp. 807--814.

\bibitem{Girshick_2015_ICCV}
R.~Girshick, ``Fast r-cnn,'' in \emph{Proceedings of the IEEE International
  Conference on Computer Vision}, 2015, pp. 1440--1448.

\bibitem{liu2019griddehazenet}
X.~Liu, Y.~Ma, Z.~Shi, and J.~Chen, ``Griddehazenet: Attention-based
  multi-scale network for image dehazing,'' in \emph{Proceedings of the IEEE
  International Conference on Computer Vision}, 2019, pp. 7314--7323.

\bibitem{li2019benchmarking}
B.~Li, W.~Ren, D.~Fu, D.~Tao, D.~Feng, W.~Zeng, and Z.~Wang, ``Benchmarking
  single-image dehazing and beyond,'' \emph{IEEE Transactions on Image
  Processing}, vol.~28, no.~1, pp. 492--505, 2019.

\bibitem{kingma2014adam}
D.~P. Kingma and J.~Ba, ``Adam: A method for stochastic optimization,''
  \emph{arXiv preprint arXiv:1412.6980}, 2014.

\end{thebibliography}

\end{document}